\documentclass[journal]{vgtc}              

\DeclareGraphicsExtensions{.png,PNG,.pdf,.PDF}

\onlineid{5777}

\vgtccategory{Research}

\def \sys {{DataCrafter}}

\def \etal {{\emph{et al}.\thinspace}}

\def \eg {{\emph{e.g}.}}
\def \ie {{\emph{i.e}.}}

\title{Human-Guided Image Generation for \\ Expanding Small-Scale Training Image Datasets}

\author{%
  Changjian Chen,
  Fei Lv,
  Yalong Guan,
  Pengcheng Wang,
  Shengjie Yu,
  Yifan Zhang,
  and Zhuo Tang
}

\authorfooter{
  \item
  	C.~Chen, F.~Lv, Y.~Guan, P.~Wang, S.~Yu, and Z.~Tang are with the College of Computer Science and Electronic Engineering, Hunan University. \\E-mail: \{changjianchen, feilv, guanyalong, wangpengcheng, ysjfloat, ztang\}@hnu.edu.cn. C.~Chen and F.~Lv are joint first authors. Z.~Tang is the corresponding author.
  \item Y.~Zhang is with the National University of Singapore. E-mail: yifan.zhang@u.nus.edu.
}

\abstract{
The performance of computer vision models in certain real-world applications (\eg, \changjian{rare wildlife observation}) is limited by the small number of available images.
Expanding datasets using pre-trained generative models is an effective way to address this limitation. 
However, since the automatic generation process is uncontrollable, the generated images are usually limited in diversity, and some of them are undesired.
In this paper, we propose a human-guided image generation method for more controllable dataset expansion.
We develop a multi-modal projection method with theoretical guarantees to facilitate the exploration of both the original and generated images.
Based on the exploration, users refine the prompts and re-generate images for better performance.
Since directly refining the prompts is challenging for novice users, we develop a sample-level prompt refinement method to make it easier.
With this method, users only need to provide sample-level feedback (\eg, which samples are undesired) to obtain better prompts.
The effectiveness of our method is demonstrated through the quantitative evaluation of the multi-modal projection method, improved model performance in the case study for both classification and object detection tasks, and positive feedback from the experts.
}


\teaser{
  \centering
  \includegraphics[width=1.02\linewidth]{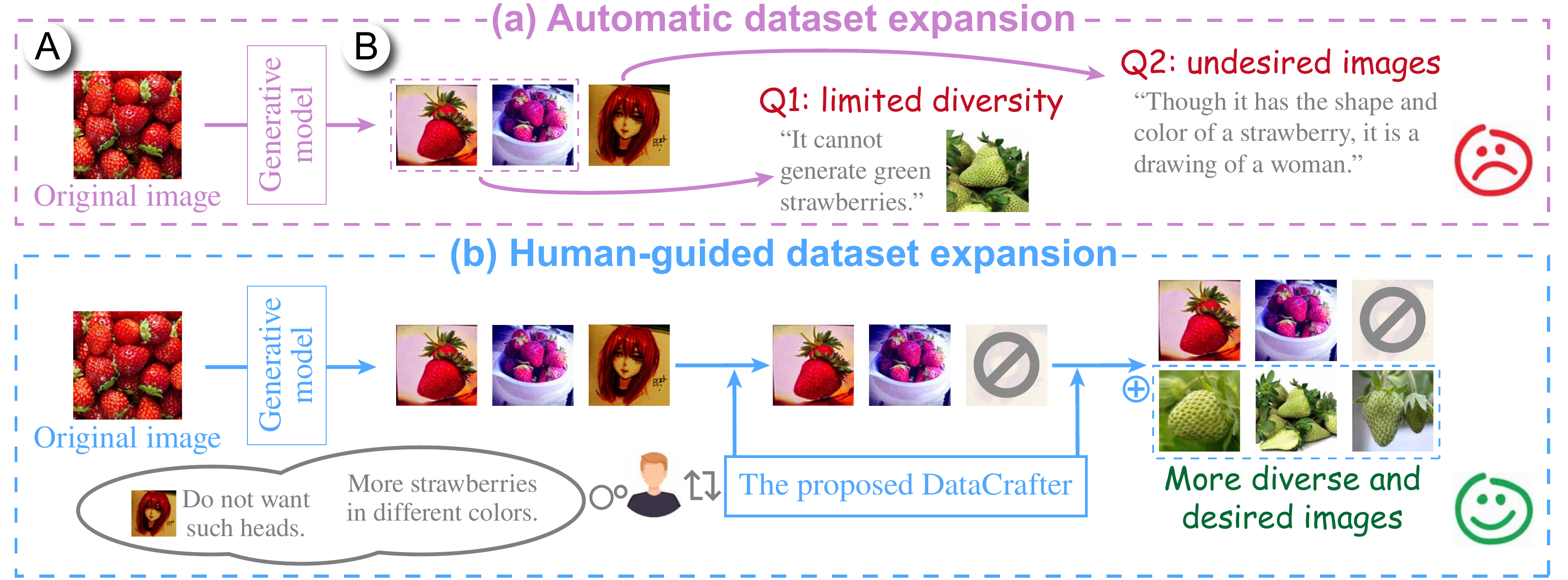}
  \caption{The comparison between the automatic and the proposed human-guided dataset expansion methods. (a) The automatic dataset expansion method is uncontrollable, which leads to limited diversity and undesired images. (b) Our proposed human-guided dataset expansion method allows users to have greater control over the generation process, resulting in more diverse and desirable images.\looseness=-1
  }
  \label{fig:teaser}
}

\graphicspath{{figs/}{figures/}{pictures/}{images/}{./}} 

\usepackage{multirow}
\usepackage{color}
\usepackage{lipsum}                    
\usepackage{amssymb,amsmath}
\usepackage{bbm} 
\usepackage{tabularx} 
\usepackage{wrapfig}
\usepackage{booktabs}
\usepackage{makecell}
\usepackage{dblfloatfix}

\newcommand{\myparagraph}[1]{\vspace{1mm}\noindent\textbf{#1}}

\newcommand{\feilvvl}[1]{\textcolor{black}{#1}}
\newcommand{\changjian}[1]{\textcolor{black}{#1}}

\usepackage{mathptmx}                  

\begin{document}


\firstsection{Introduction}

\maketitle
\fontsize{9}{9} 

The success of computer vision relies on a large number of images~\cite{yang2023foundation}.
However, collecting such large image datasets is expensive and time-consuming~\cite{chen2022towards}.
Moreover, in certain real-world applications (\eg, \changjian{rare wildlife observation}), it is impractical to gather a large number of images due to factors like the rarity of \changjian{wildlife}~\cite{hasani2022artificial}.
Recently, due to the advancements in pre-trained text-to-image generative models (\eg, Stable Diffusion~\cite{rombach2022high}), \textbf{dataset expansion} has emerged as an effective way to address this issue~\cite{zhang2023expanding}.
Instead of generating images freely with generative models, dataset expansion initially utilizes several original images (\eg, Fig.~\ref{fig:teaser}A) as references and text prompts as perturbations to generate several similar but not identical images (\eg, Fig.~\ref{fig:teaser}B).
Such a strategy ensures that the generated images are task-relevant, which improves model performance.
Then, these generated images, along with the original ones, are utilized to train models for downstream tasks, such as classification.

\changjian{Although expanding the dataset is effective for generating training images, their benefits diminish, and in some cases, they may even harm model performance as more images are generated~\cite{yamaguchi2023on}.
This is primarily due to two quality issues in them (Fig.~\ref{fig:teaser}(a)).}
First, dataset expansion often falls short in promoting the diversity of the generated images.
For example, generating images based on the original image in Fig.~\ref{fig:teaser}A results in only red strawberries,
whereas green strawberries, which are rare but do exist in the real world, are not generated (Fig.~\ref{fig:teaser}Q1).
Second, due to the ambiguity of natural language prompts, the generated images are not always of interest. 
For example, when generating strawberries, a drawing of a woman is generated because its shape and color are similar to a red strawberry (Fig.~\ref{fig:teaser}Q2).

An effective way to address these quality issues is by allowing humans to guide the generation process through refining prompts based on their knowledge~\cite{dunlap2023diversify}. 
However, injecting human guidance into the generation process faces two challenges.
First, the content of the generated images needs to be examined to identify missing or undesired ones.
This process can be tedious and time-consuming, as the content of each image needs to be inspected individually.
Therefore, a tool that facilitates efficient exploration is needed.
Second, due to the complexity and ambiguity of natural language, it is challenging for humans to refine prompts when a lack of diversity or undesired images are identified.

To address these challenges, we develop {\sys}, a visual analysis tool to help 1) explore the original and generated images efficiently and 
2) \changjian{provide sample-level feedback to} refine the prompts for better training image datasets.
Given a set of original images and prompts, several images are generated.
To help better explore the original and generated images, inspired by a recent work~\cite{li2024visual}, we use multi-modal language models (\eg, GPT-4~\cite{openai2023gpt4}) to extract descriptive labels for their content.
These content labels make it easier to access the content of the images rather than examining them individually.
With these content labels, a multi-modal projection method is developed to project both the images and content labels in a two-dimensional plane for exploration. 
\feilvvl{Although MFM~\cite{ye2024mfm} has been proposed recently for multi-modal projection}, 
we theoretically and experimentally validate that it performs well in a \textit{many-to-one} setting, \changjian{where each image has only a single label, but falls short in a \textit{many-to-many} setting, where images can have multiple labels -- a scenario that is more common in practice.}
Based on the theoretical analysis, we propose a contrastive-learning-based algorithm to better preserve both inter- and intra-modality relationships.
With the multi-modal projection result, users explore and analyze the original and generated images, and refine prompts to generate more diverse and desired images.
\changjian{Through literature reviews and expert feedback, we find that} it is easier for users to provide sample-level feedback (\eg, which samples are undesired) than directly editing the prompts.
Therefore, we develop a sample-level prompt refinement method, which recommends better prompts based on the sample-level feedback.
The effectiveness of our method is demonstrated through a quantitative evaluation of the multi-modal projection method, improved model performance in the case study for both classification and object detection tasks, and positive feedback from experts.
\feilvvl{The source code is available at: https://github.com/hnu-vis/DataCrafter.}



In summary, the contributions of this work include:

\begin{itemize}[nosep]
\item\noindent{\textbf{A multi-modal projection method} to preserve both inter- and intra-modality relationships, which is theoretically and experimentally better than the existing methods.} 
\item\noindent{\textbf{A sample-level prompt refinement method} that lowers the barrier of refining prompts for dataset expansion.}
\item\noindent{\textbf{A visual analysis tool} that supports human-guided image generation for better training image datasets.}

\end{itemize}



\section{Related Work}

Our work focuses on the interactive generation of training image datasets.
Existing research in this area can be categorized into interactive label generation and interactive image generation.
\looseness=-1

\subsection{Interactive Label Generation}
Interactive label generation aims to facilitate the labeling process while minimizing human effort.
Based on how labels are obtained, these methods can be classified into two categories: propagation-based methods and function-based methods.

\textbf{Propagation-based methods} first allow users to label a few samples.
These labels are then propagated to other unlabeled samples to obtain more labels. 
The \feilvvl{“inter-active learning”} framework \changjian{is} introduced by H{\"o}ferlin~\etal~\cite{hoferlin2012inter} to obtain labels for the classification task.
This framework allowed users to either label samples recommended by an active learning model or select informative samples for labeling through visualizations.
These labels were propagated to other unlabeled samples to obtain more labels, which were used to further enhance the underlying model. 
This strategy was also substantiated by other work for obtaining labels for the classification task, such as DataLinker~\cite{chen2021interactive} and FSLDiagnotor~\cite{yang2022diagnosing}.
Apart from the classification task, other tasks were also studied. 
For the object detection task, 
MutualDetector~\cite{chen2022towards} utilized a multi-type relational co-clustering algorithm to cluster samples and labels simultaneously, \changjian{which facilitates the exploration and annotations of bounding boxes and labels}.
For the temporal action localization task, 
\changjian{Chen~\etal~\cite{chen2024enhancing} developed a storyline visualization to display the actions and their alignments, which helps users} correct wrong localization results and misalignments for propagation.

\textbf{Function-based methods} enable users to define functions to generate labels instead of directly labeling samples.
Hoque~\etal~\cite{hoque2023visual} introduced the Visual Concept Programming,
which utilized a self-supervised learning approach to extract visual concepts (\eg, objects and parts) from images.
Users explored these visual concepts with the help of visualizations and composed labeling functions (\eg, Concept$_\text{Transportation}$ +above + Concept$_\text{Water}$ =Class$_\text{Boat}$) with them to generate labels.
He~\etal~\cite{he2024videopro} extended this idea to generate labels for video data.
To ensure the quality of labels, visual concept programming usually iterates for many rounds, which makes it challenging for users to understand the process.
To tackle this issue, Li~\etal~\cite{li2024evovis} proposed EvoVis, which combined relationship analysis and temporal overview to help users explore the visual concept programming iterations. 

Though effective, these methods require images to generate labels for computer vision tasks.
However, in many real-world applications (\eg, \changjian{rare wildlife observation}), it is infeasible to acquire a large number of images. 
To tackle this, we combine dataset expansion and visualization to help generate training images to improve model performance.

\addtocounter{figure}{1}
\begin{figure*}[b]
    \centering
    \includegraphics[width=\linewidth]{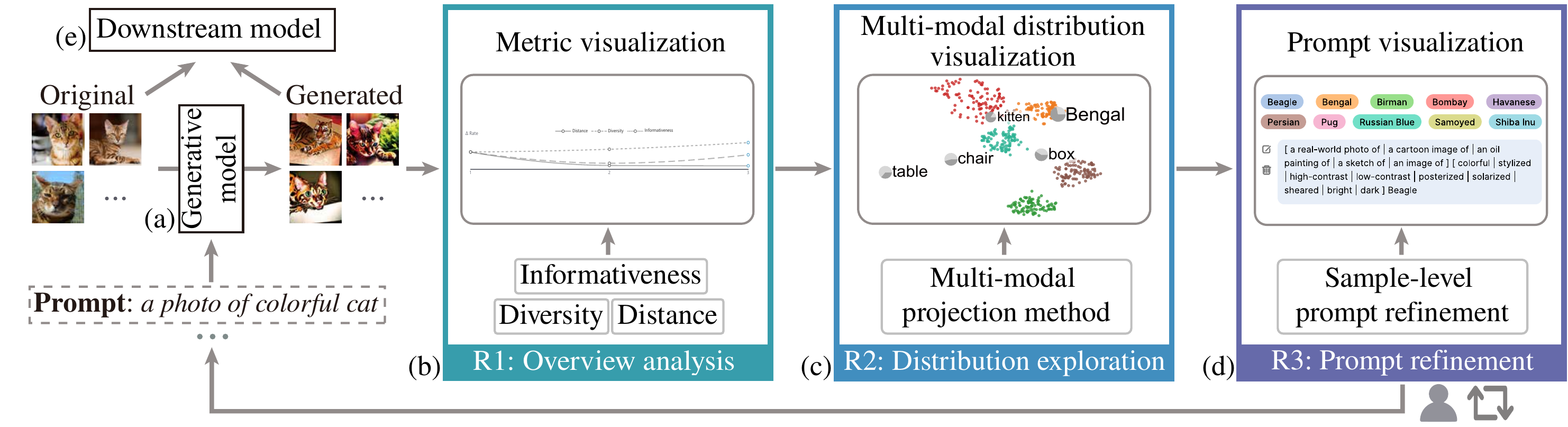}
    \caption{Method overview: (a) given a small number of original images and prompts, a set of images is generated; (b)-(d) three visualizations to help explore the original and generated images and refine the prompt interactively; (e) the original and generated images are combined to train the downstream model for better performance.}
    \label{fig:pipeline}
\end{figure*}

\addtocounter{figure}{-2}
\begin{figure}[t]
    \centering
    \includegraphics[width=\linewidth]{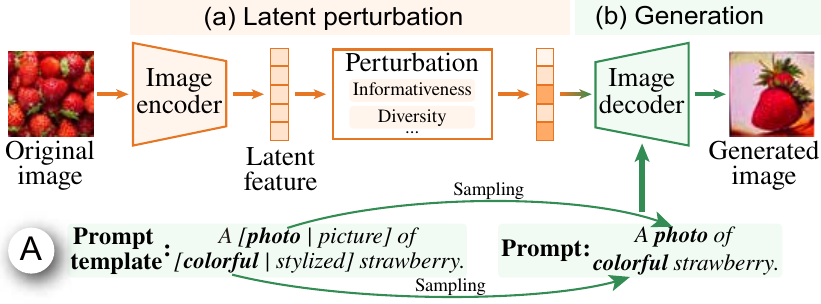}
    \caption{Dataset expansion consists of two main steps: (a) latent perturbation and (b) image generation.}
    \label{fig:background}
    \vspace{-2mm}
\end{figure}
\addtocounter{figure}{1}

\subsection{Interactive Image Generation}
Depending on whether the downstream tasks are taken into account during the generation process, \changjian{interactive image generation} methods can be categorized into task-independent and task-dependent methods.

\textbf{Task-independent methods} aim to help users generate images that satisfy their personal requirements.
\changjian{Most of them focus on prompt engineering for image generative models.}
PromptMagician~\cite{feng2024promptmagician} developed a multi-level visualization for exploring generated images and combined it with a prompt database to recommend keywords for text prompt refinement.
This exploration and recommendation strategy was also used by Reprompt~\cite{wang2023reprompt}, PromptCharm~\cite{wang2024promptcharm}, and Promptify~\cite{brade2023promptify}.
For concepts that were challenging to convey through text (\eg, the color style of an image),
Chung~\etal~\cite{chung2023promptpaint} developed PromptPaint to help create prompts beyond text.
PromptPaint enabled users to overlay various colors or apply different colors to distinct areas of an image.
This feedback was mixed with text prompts to refine the generated images.
As prompt refinement often requires many trials and errors, the prompt history is rich with valuable information that can offer users insights into past explorations and illustrate how changes to the prompt influence the generated image.
In this regard, Guo~\etal~\cite{guo2024prompTHis} introduced the Image Variant Graph, facilitating the comparison of prompt-image pairs and exploration of editing histories.

All the methods mentioned above focus on generating images that meet users' personal requirements without considering the downstream tasks.
Directly applying them to expanding datasets will result in many task-irrelevant images, which is not helpful or even harmful to model training~\cite{xu2021dash}.
Therefore, task-dependent methods are proposed.

\textbf{Task-dependent methods} consider the needs of downstream tasks during the generation process to enhance model performance.
The very first method of this kind, VATLD, was introduced by Gou~\etal~\cite{gou2020vatld} to expand training datasets for object detection. 
VATLD adapted a representation learning method to extract data semantics and help users explore them in multiple coordinated visualizations.
Once a certain type of object fails to be detected, users can generate more such objects with the help of the representation learning method.
He~\etal~\cite{he2021can} generalized this idea to semantic segmentation by introducing a context-aware spatial representation learning method, 
\feilvvl{which captures key spatial details about objects, including position, size, and aspect ratio to help assess model performance and generate new test cases for various driving scenes.}

Although these task-dependent methods can effectively expand training datasets to enhance the model performance, they require a large number of images to train the representation learning models. 
However, such a large number of images are often unavailable.
To tackle this issue, we utilize pre-trained generative models (\eg, Stable Diffusion~\cite{rombach2022high}) to generate images.
To facilitate the generation, we develop a multi-modal projection method to help explore the original and generated images, and a sample-level prompt refinement method to help refine the prompts easily.


\section{Background: Data Expansion}
\label{sec:background}

Recent research has proposed various data expansion methods~\cite{zhu2024distribution,li2024semantic,dunlap2023diversify,trabucco2023effective,zhang2023expanding}. These methods can all be summarized into a common pipeline, as illustrated in Fig.~\ref{fig:background}. 
This pipeline consists of two main steps: latent perturbation and image generation.

\myparagraph{Latent perturbation}.
Given an original image, a latent feature is extracted through the image encoder of a generative model.
Then, the latent feature is perturbed to encourage similar but not identical images to be generated in the following step.
To ensure the generated images benefit the downstream tasks, constraints are introduced in the perturbation process.
These constraints optimize certain metrics to ensure the quality and task relevance of the generated images. 
Typical metrics~\cite{zhang2023expanding,zhu2024distribution} include:
1) \emph{Informativeness}.
According to the study of Zhang~\etal~\cite{zhang2023expanding}, higher task-related informativeness provides more detailed information about the downstream task, which promotes model performance.
2) \emph{Diversity}.
As previous studies~\cite{chen2020oodanalyzer, shi2016diversifying} pointed out, the diversity of training datasets is essential for the generalization ability of models.
A higher diversity usually leads to better performance.



\begin{figure*}[b]
    \centering
    \includegraphics[width=\linewidth]{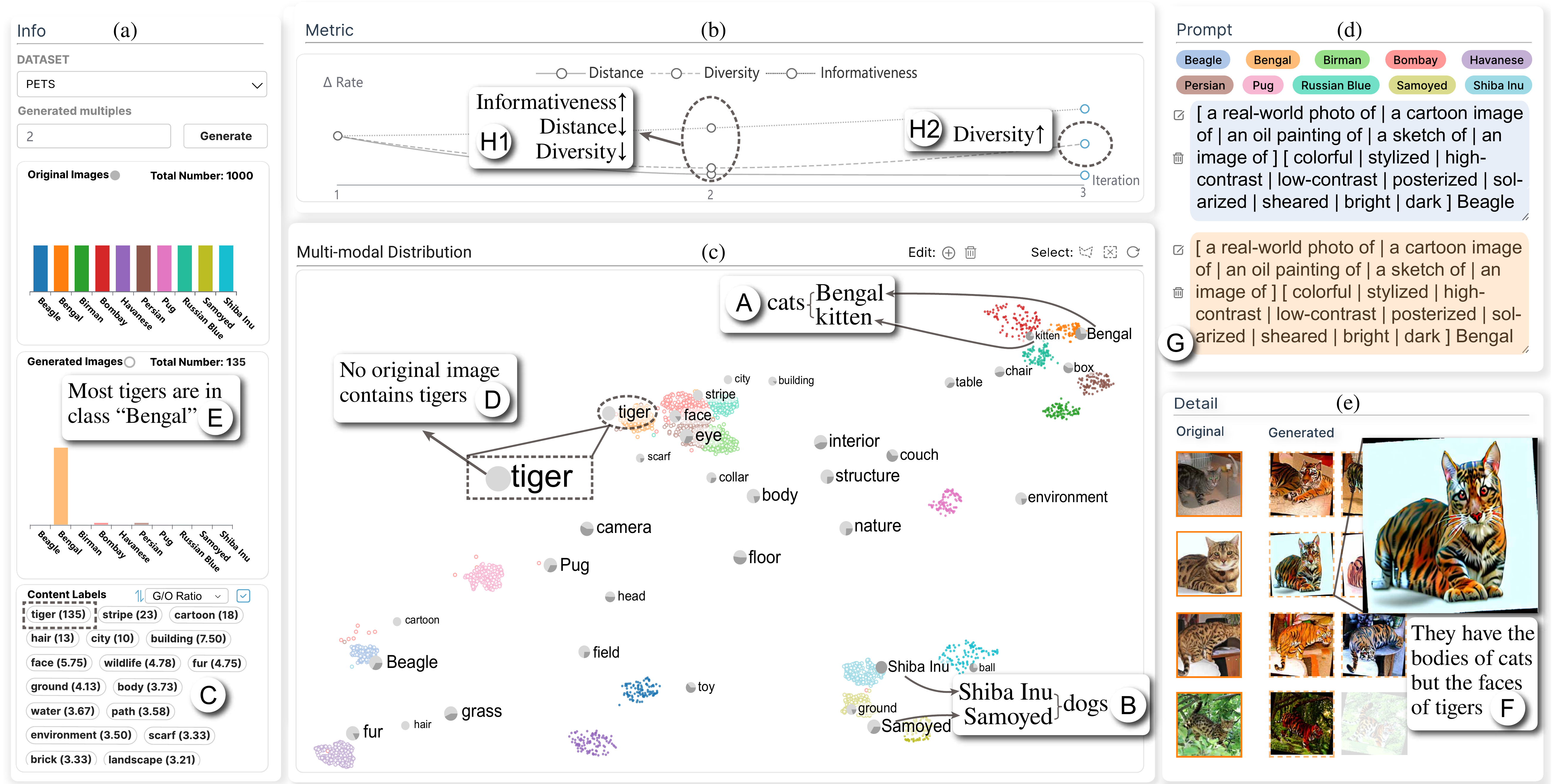}
    \caption{{\sys}: (a) an information panel to show the numbers of original and generated images and content labels; (b) a metric visualization to show important metrics of the generated images; (c) a multi-modal distribution visualization to show images in the context of content labels; (d) a prompt visualization to show prompts; (e) a detail panel to show selected images.}
    \label{fig:system}
\end{figure*}

\myparagraph{Image generation}.
After perturbing the latent features, these features, combined with the text prompt templates, are fed into the image decoder of the generative model to generate new images.
Prompt templates instead of fixed prompts are used because they enhance the diversity of the generated images and are widely adopted in existing methods~\cite{li2024semantic, dunlap2023diversify, trabucco2023effective}.
A prompt template offers several options (\eg, "[photo | picture]" in Fig.~\ref{fig:background}A) for specific parts of the prompts. 
Each time an image is generated, a prompt is sampled from the template.
Due to the poor readability of prompt templates, we use prompts to illustrate the basic ideas of our method in Secs.~\ref{sec:requirement} and~\ref{sec:method}.



\section{Requirement Analysis}
\label{sec:requirement}


This work was developed in close collaboration with four machine learning experts (E1-E4). 
E1 and E2 are Ph.D. students.
E1 has published a pioneering paper on data expansion.
Both E1 and E2 are highly knowledgeable about the latest data expansion methods and are keen to push the boundaries of this field by uncovering new insights.
E3 is a researcher in an \feilvvl{academic institute}, and E4 is a machine learning engineer in a technology company.
Both E3 and R4 are actively involved in deploying data expansion methods in real-world applications, such as anomaly detection. 
During deployment, they observed that many generated images did not meet their expectations. 
Consequently, they expressed a need for more controllable data expansion.
We gathered the following requirements from four semi-structured interviews with \changjian{the} experts, each lasting between 40 and 70 minutes.

\noindent\textbf{R1: Examining the overview of the generated images and their dynamic changes through the generation process}.
All the experts expressed the need to have an overview of the generated images.
``As the images are generated gradually instead of all at once, I need to understand what is happening during the generation process. This way, I can decide the right moment to interrupt the process.'' E3 said.
E2 also pointed out that the overview should be simple and intuitive.
It should clearly show how the generated images are evolving over time, allowing users to easily track the progress.

\noindent\textbf{R2: Exploring the original and generated images efficiently}.
To understand how the generated images impact the downstream model performance, 
it is necessary to analyze and compare them with the original images.
 \changjian{The experts usually use grid visualizations or scatterplots to explore images.
Grid visualizations place images into grid cells, making them convenient for examining image content.
However, the experts found grid visualizations were insufficient for revealing cluster separation, which is essential to identify undesired generated images.
Scatterplots are effective in comparing the distributions of the original and generated images and conveying clusters.}
However, the experts pointed out that identifying quality issues in the generated images with \changjian{scatterplots} requires examining the content of each image individually, which can be tedious.
Therefore, an efficient way to explore the original and generated images is required.

\noindent\textbf{R3: Refining prompts effectively and efficiently to guide the generation process}.
When the generated images are unsatisfactory, modifying the prompts is an effective way to guide the generation process~\cite{feng2024promptmagician}.
However, manually refining the prompts is tedious and time-consuming, especially for novice users.
\changjian{Although some methods have been proposed, such as prompt keyword recommendation (PromptMagician~\cite{feng2024promptmagician} and PromptCharm~\cite{wang2024promptcharm}), to facilitate prompt refinement, it is still challenging because users have to edit prompts directly.}
Therefore, all the experts sought a more user-friendly and efficient method for prompt refinement.
``When I see the generated images, I can quickly tell whether they are desired or not. 
However, modifying the prompts to avoid generating undesired images is challenging because the generation models are ‘black boxes,’ requiring multiple attempts.'' E4 said.


\section{{\sys} Visualization}
\label{sec:method}

Based on the requirements, we develop {\sys} to facilitate human-guided image generation for dataset expansion.
Fig.~\ref{fig:pipeline} shows its overview.
Given a small number of original images and some text prompts,
a set of images is generated (Fig.~\ref{fig:pipeline}(a)).
Users can have an overview of the generated images in the metric visualization (Fig.~\ref{fig:pipeline}(b), \textbf{R1}) and explore them along with the original images in the multi-modal distribution visualization (Fig.~\ref{fig:pipeline}(c), \textbf{R2}).
During the exploration, users can refine the prompts (Fig.~\ref{fig:pipeline}(d), \textbf{R3}) by providing sample-level feedback, such as removing some undesired images.
The refined prompts are fed into the generative model to generate more images.
This process iterates several times until the generated images are satisfying.
Then, the original and the generated images are combined to train the downstream model for better performance (Fig.~\ref{fig:pipeline}(e)).

\subsection{Metric Visualization}
The metric visualization gives an overview of the generated images.
The existing methods~\cite{zhang2023expanding,zhu2024distribution} optimize certain metrics in the latent spaces to ensure the quality and relevance of the generated images.
This inspires us to use these metrics to give an overview of the generated images.
As shown in Fig.~\ref{fig:system}(b), the metrics of the generated images over iterations are shown as a line chart.
To determine which metrics to include, we systematically review the dataset expansion papers and collect the metrics used in them.
The detailed review results can be found in the supplemental material.
According to this review, we identify three metrics: informativeness, diversity, and distance.

\myparagraph{Informativeness}.
As pointed out in Sec.~\ref{sec:background},
high task-related informativeness promotes model performance.
According to Zhang~\etal~\cite{zhang2023expanding}, an image with high task-related informativeness should have high prediction entropy while maintaining the class semantic.
Thus, the informativeness of a generated image is measured by
\begin{equation}
    \text{S}_{inf} = \mathrm{Entropy}(p^{\prime}) + p^{\prime}_j, \quad \mathrm{s.t.}\ j =\mathrm{argmax}(p).
\end{equation}
The first term measures the prediction entropy, and the second term measures how well the generated image maintains the class semantics of the original image. 
$p$ and $p^{\prime}$ are the prediction distributions of the original and generated images, respectively.
$p_j^{\prime}$ is the probability that this generated image belongs to $j$-th class.
The prediction distributions can be obtained with zero-shot classification models, such as CLIP~\cite{radford2021learning}.
The informativeness of a set of generated images is calculated by averaging the informativeness of each individual image.



\myparagraph{Diversity}.
A common way to measure the diversity of a set of images is by averaging the distances between each image to their centers~\cite{tang2006analysis}:
\begin{equation}
\label{eq:div}
    \text{S}_{div} =\frac{1}{M} \sum_i^M D(v_i^{\prime},\bar{v^{\prime}}).
\end{equation}
Following the work of Zhang~\etal~\cite{zhang2023expanding}, we use the Kullback–Leibler (KL) divergence as the distance measure ($D(\cdot, \cdot)$).
$M$ is the number of generated images.
$v_i^{\prime}$ is the feature vector of the $i$-th generated image, which is extracted with CLIP.
$\bar{v^{\prime}}$ is the mean feature vector of the generated images.
Since the within-class diversity is more useful than the diversity over the whole dataset in practice~\cite{liu2022memory}, we calculate the diversities for each class and average them to obtain the final one.



\myparagraph{Distance}.
The distance between the original and generated images is widely used to measure the quality of generated images.
If the distance is too high, the generated images differ \changjian{greatly} from the original ones, hindering the training process. 
Conversely, if the distance is too low, the generated images are nearly identical to the original ones, offering no benefit.
Thus, achieving an appropriate level of distance is crucial~\cite{zhang2023expanding}.
A recent study~\cite{jayasumana2024rethinking} shows that the CLIP Maximum Mean Discrepancy (CMMD) measure is better than traditional ones (such as Frechet Inception Distance~\cite{heusel2017gans}).
Therefore, we utilize CMMD to measure the distance between the original and generated images:
\begin{equation}
    \begin{aligned}
        \text{S}_{dis}^{2} = & \frac{1}{N(N-1)} \sum_{i=1}^{N} \sum_{j \neq i}^{N} K\left(v_{i}, v_{j}\right) -\frac{2}{N M} \sum_{i=1}^{N} \sum_{j=1}^{M} K\left(v_{i}, v^{\prime}_{j}\right) \\
        & +\frac{1}{M(M-1)} \sum_{i=1}^{M} \sum_{j \neq i}^{M} K\left(v^{\prime}_{i}, v^{\prime}_{j}\right).
    \end{aligned}
\end{equation}
Here, $N$ is the original image number.
$K(x,y)=\mathrm{exp}(-{||x-y||}^2 /2{\sigma}^2)$ is the Gaussian kernel with a variance of $\sigma$.

\subsection{Multi-modal Distribution Visualization}
A core requirement from the experts is the efficient exploration of the original and generated images (\textbf{R2}).
\changjian{In the literature, many visualization methods have been proposed for image exploration, which can be broadly classified into grid visualizations~\cite{zhou2024cluster, chen2024unified, chen2020oodanalyzer} and scatterplots~\cite{xie2019semantic, chen2021interactive, liu2019interative, jia2022towards}. 
Grid visualizations are convenient for examining image content but fall short in revealing cluster separation~\cite{micallef2017towards}, which is essential for identifying undesired images in generated images.
In contrast, scatterplot-based methods are effective at cluster separation and analyzing the data distributions~\cite{xiang2019interactive}.
However, assessing quality issues} using \changjian{scatterplots} typically requires examining each image individually, which can be tedious and time-consuming.


To tackle these issues, inspired by a recent work~\cite{li2024visual}, we utilize multi-modal language models (\eg, GPT-4) to extract descriptive labels from images.
These labels, referred to as content labels, describe the content of the images.
By examining these content labels, users can quickly discern the contents of the images without needing to examine each one individually.
Based on this idea, we design the multi-modal distribution visualization, which overlays the content labels on the scatterplot of images to allow users to explore the original and generated images and quickly access their content.

The multi-modal distribution visualization is shown in Fig.~\ref{fig:system}(c).
Each original (generated) image is presented as a $\vcenter{\hbox{\includegraphics[height=1.5\fontcharht\font`\B]{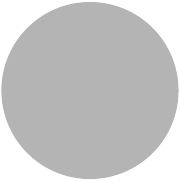}}}$ ($\vcenter{\hbox{\includegraphics[height=1.5\fontcharht\font`\B]{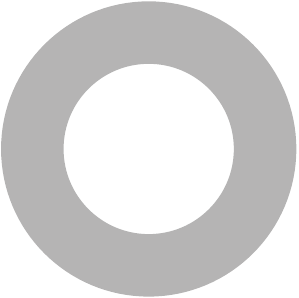}}}$).
Their colors encode the classes.
The content labels are placed near their images.
The pie chart ($\vcenter{\hbox{\includegraphics[height=1.5\fontcharht\font`\B]{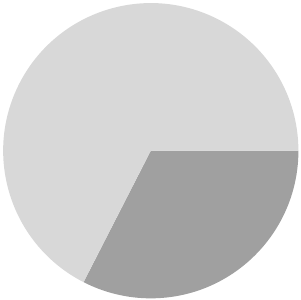}}}$) before each content label shows the percentage of original and generated images that contain this label.
The size of the pie chart represents the total number of images that contain this label.
To ensure that the content labels accurately reflect the context of the images, we need to 1) group similar images together, 2) match images with their corresponding content labels, and 3) group similar content labels together. 
To achieve this, we propose a multi-modal projection method called M2M. 
Based on M2M, we develop a multi-level exploration strategy to handle the large number of images and content labels effectively.


\subsubsection{Multi-modal projection}
\myparagraph{Problem setting}. 
Given $N_I$ images $X = \{I_i\}_{i=1}^{N_I}$ and $N_L$ content labels $Y = \{L_i\}_{i=1}^{N_L}$, each \feilvvl{image contains} one or more content labels.
They can be modeled as a bipartite graph $(X, Y, E)$.
In $E$, each element $E_{ij}$ represents image $I_i$ \feilvvl{contains content} label $L_j$.
Its weight represents the distance between $I_i$ and $L_j$ in the high-dimensional space (\eg, the CLIP embedding space).
Given the bipartite graph, the multi-modal projection aims to project both the images and content labels into a 2D plane such that similar images (content labels) are placed together, and images are paired with their corresponding content labels. 
Since the number of content labels in an image affects the complexity of the problem, we categorize the problem into two settings:
\begin{itemize}[nosep]
\item\noindent{\textit{Many-to-one} setting: an image contains no more than one label.} 
\item\noindent{\textit{Many-to-many} setting: an image can contain more than one label.}
\end{itemize}

\myparagraph{Existing method for the many-to-one setting}. 
Recently, Ye~\etal~\cite{ye2024mfm} proposed a multi-modal projection method, MFM, and achieved remarkable results.
Although MFM was theoretically applicable in the \textit{many-to-many} setting, Ye~\etal~\cite{ye2024mfm} mainly validated its effectiveness in the \textit{many-to-one} setting.
\feilvvl{If we} apply MFM to the \textit{many-to-many} setting, the images of different classes are cluttered together, and the similarities between images and labels are not preserved well (\eg, Fig.~\ref{fig:projection}(c)).
\changjian{This} is caused by the multi-modal distance order loss used in MFM. 
The multi-modal distance order loss ensures the relative distance orders of images to content labels:
\begin{equation}
    \begin{aligned}
    \frac{-\sum_{i, j<t}\pi((h(I_j,L_i) - h(I_t,L_i))\times (l(I_j,L_i) - l(I_t,L_i)))}{\sum_{i,j}l(I_j,L_i)^2}.
    \end{aligned}
\end{equation}
Here $h(\cdot, \cdot)$ and $l(\cdot, \cdot)$ are the distances in the high-dimensional and low-dimensional spaces, respectively.
$\pi(x)=0$ if $x>0$; otherwise $\pi(x)=-x$.
If the relative distance orders of images to content labels are exactly preserved, the multi-modal distance order loss will be zero.

In the \textit{many-to-one} setting, we can prove the following theorem.
\newtheorem{theorem}{Theorem}
\begin{theorem}
Given a bipartite graph (X, Y, E) in the many-to-one setting where each point in X is connected to at most one point in Y, there exist mappings that project points in X and Y in a 2D plane such that the distance orders of the neighbors of points in Y can be exactly preserved. 
\end{theorem}
A formal proof of the correctness of this theorem is given in the supplemental material.
This theorem implies that the distance orders of images to the content labels can be exactly preserved (\ie, the multi-modal distance order loss is zero).
It explains why the multi-modal distance order loss works well in the \textit{many-to-one} setting.

However, in the more common \textit{many-to-many} setting, the problem becomes more complex.
In this setting, we have the following theorem.
\begin{theorem}
Given two point sets X and Y in a 2D plane, the number of different orders of points in X to points in Y is no more than
$$
n(n-1)(n^2-n+2)/8 + 1,
$$
where $n=|X|.$
\end{theorem}
Given this theorem, we have the following lemma.
\newtheorem{lemma}[theorem]{Lemma}
\begin{lemma}
Given a bipartite graph (X, Y, E) in a many-to-many setting where each point in X can be connected to more than one point in Y, there does not always exist one mapping that projects points in X and Y in a 2D plane such that the distance orders of the neighbors of points in Y can be exactly preserved if 
$$
n(n-1)(n^2-n+2)/8 + 1< n!,
$$
where $n=|X|.$
\end{lemma}
The formal proofs of the theorem and lemma can be found in the supplemental material.  
The equation in the lemma holds when $n > 3$.
It implies that the distance orders of images to the content labels cannot always be exactly preserved when the number of images is larger than three.
This explains why the multi-modal distance order loss does not work well in the \textit{many-to-many} setting.

\myparagraph{Our method for the many-to-many setting}.
To find more suitable losses for the many-to-many setting, we get inspiration from multi-modal learning.
In this field, constrastive loss has become one of the most popular methods for multi-modal learning tasks.
And its effectiveness is widely validated.
Moreover, a recent theoretical work~\cite{wen2024convergence} has proven that an approximate stationary solution can be obtained for contrastive learning. 
Since the contrastive loss is suitable for multi-modal tasks in both practice and theory, we develop a conservative learning-based multi-modal projection method.

Our method utilizes a six-layer feed-forward neural network for projection.
With this network, our method aims to minimize the weighted sum of three contrastive losses:
\begin{equation}
\begin{aligned}
    CL_{II} + \omega_1 CL_{IL} + \omega_2 CL_{LL}.
\end{aligned}
\end{equation}
The first and third terms ensure similar images and content labels are placed together.
The second term encourages images to be close to their corresponding content labels.
$\omega_1$ and $\omega_2$ are weights to balance these terms and are determined by the multi-task learning method~\cite{Liu_Liang_Gitter_2019}.
All three terms share a common contrastive loss.
Such a loss for a positive pair ($x_i$, $x_j$) is defined as 
\begin{equation}
    -\log \frac{\ \exp(\mathrm{sim}(x_i, x_j) / \tau)}{\sum_{t=1}^{2B} \mathbbm{1}_{[t \neq i]} \ \exp(\mathrm{sim}(x_i, x_t) / \tau)}.
\end{equation}
$B$ represents the size of a mini-batch during training.
$\mathbbm{1}_{[t \neq i]}$ is \feilvvl{an indicator} function.
$\mathbbm{1}_{[t \neq i]}=1$ if $t \neq i$; otherwise $\mathbbm{1}_{[t \neq i]}=0$.
$\mathrm{sim}(\cdot, \cdot)$ is the similarity between two points.
The difference between the three terms lies in the definition of positive and negative pairs.

For the \underline{\normalsize first term},
each neighbor among the $k$-nearest neighbors of a given image, along with the given image, is considered a positive image pair. 
Conversely, each image outside the $k$-nearest neighbors of the given image, along with the given image, is considered a negative image pair.
The positive and negative label pairs in the \underline{\normalsize third term} are constructed in a similar way.
For the \underline{\normalsize second term}, an image and one of its content labels are considered as a positive pair. 
For the negative pair, a straightforward way is to pair an image and a label that does not appear in that image.
However, we experimentally find that such a strategy leads to the content labels collapsing into a small area.
The underlying reason is the uneven frequency distribution of labels, which follows the "80-20" rule: 80\% of the labels appear only 20\% of the time.
Therefore, these rare labels are more likely used for constructing negative pairs.
Such negative pairs do not provide much information because an image is not likely to contain rare labels.
To this end, we utilize a frequency-biased sampling strategy to construct the negative pairs.
Given an image, we sample several labels that do not appear in that image with probabilities \{$f_i$\}.
$f_i$ is the frequency of the $i$-th label over the total frequency of all labels.

\subsubsection{Hierarchical exploration}
Since the number of content labels is typically large, displaying all of them simultaneously can cause visual clutter.
To address this, we employ a tree cut algorithm to dynamically display content labels of interest~\cite{furnas1986generalized, lyu2024supercomputer, chen2021interactive}.
Specifically, we organize the content labels into a hierarchy.
The hierarchy is constructed using an agglomerative method, a widely used and computationally efficient hierarchical clustering method~\cite{reddy2018survey}.
The names of the inner nodes are generated based on their children using GPT-4.
Initially, several nodes in the hierarchy are displayed.
If users are interested in the content labels in an area, they can zoom in for more detailed content labels.
For each node $x$ in the hierarchy, its ``Degree of Interest'' (DOI) is calculated as follows:
\begin{equation}
\mathrm{DOI}(x|y)=\mathrm{API}(x)-\mathrm{TD}(x,y),
\end{equation}
where $\mathrm{API}(x)$ represents the priori importance of node $x$, determined by the ratio of generated images to the number of original images that include node $x$.
This ratio is chosen because high ratios indicate imbalances between the original and generated images, which may suggest potential quality issues.
$\mathrm{TD}(x,y)$ is the tree distance between node $x$ and the focused node $y$, which is the closest node to the center of the currently focused area.

\subsection{Interactions for Human-Guided Image Generation}

\subsubsection{Sample-Level Prompt Refinement}
To guide the generation process for more desired samples, a straightforward way is to modify the prompts~\cite{li2024semantic}.
However, directly modifying the prompts (prompt-level refinement) is a tedious trial-and-error process because of the complexity and ambiguity of natural language and the black-box nature of the generative models~\cite{feng2024promptmagician}.
The work of Endert~\etal~\cite{endert2011observation} shows that users are more comfortable interacting with what they see instead of the underlying models to refine the models.
Inspired by this work, we propose a more user-friendly refinement strategy, the sample-level prompt refinement.
In this strategy, users only need to provide feedback on the images (\eg, which images are undesired).
Then, an evolutionary-algorithm-based \changjian{strategy} is utilized to automatically refine the prompts to accommodate the feedback. 

\begin{figure}[t]
    \centering
    \includegraphics[width=\linewidth]{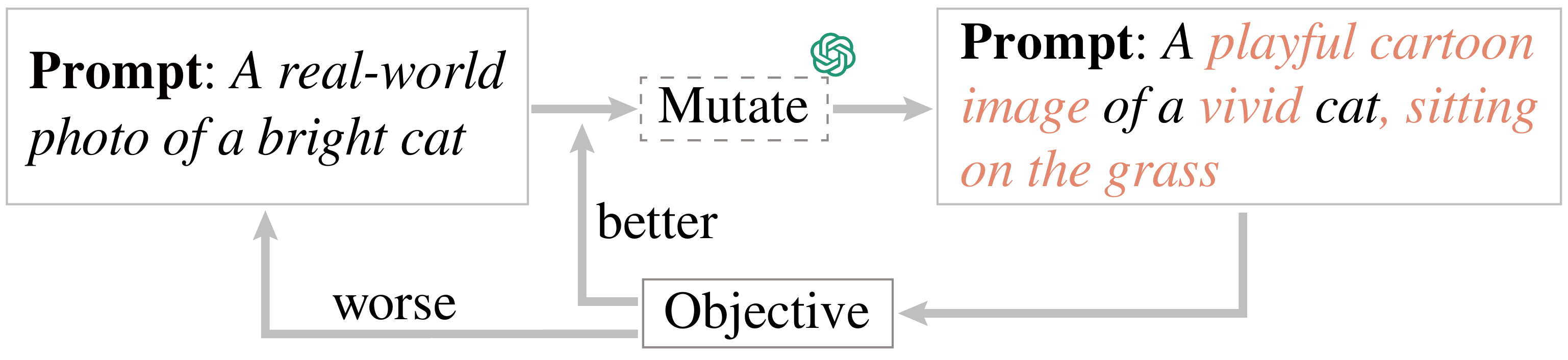}
    \caption{The evolutionary-algorithm-based prompt refinement.}
    \label{fig:evo}
\end{figure}

\myparagraph{Algorithm framework}.
To automatically refine prompts based on the feedback on generated images, 
the algorithm needs to be able to 1) handle discrete \changjian{prompt} optimization
and 2) optimize different objectives because users may provide different sample-level feedback.
Evolutionary algorithms naturally satisfy these two requirements~\cite{zhou2019evolutionary}. 
Therefore, we choose them for prompt refinement.
The pipeline of the evolutionary-algorithm-based prompt refinement \changjian{strategy} is shown in Fig.~\ref{fig:evo}.
Given a prompt, we utilize LLMs (\eg, GPT-4) to mutate it while ensuring it is coherent and human-readable.
For example, the prompt \emph{``A real-world photo of a bright cat''} is mutated to \emph{``A playful cartoon image of a vivid cat, sitting on the grass''}.
Then, the mutated prompt is evaluated on different objectives based on different feedback.
Since it is difficult to evaluate the mutated prompt directly~\cite{guoconnecting}, we generate several proxy images with the mutated prompt and use these images as proxies for evaluation. 
If the mutated prompt is better than the previous one, it is used for further mutation; otherwise, the mutated prompt is abandoned, and the previous one is preserved for further mutation.
This process iterates several times until the gain in the objective is less than a small threshold.

\myparagraph{Evaluation objective}.
Different feedback requires different evaluation objectives.
To determine how many types of feedback users may provide, we observe that the feedback on generated images can be modeled as operations on a single set. 
According to set theory~\cite{hausdorff2007set}, there are three such operations: adding, removing, and transforming elements. 
When we introduce these operations to the experts, they point out that adding and removing elements is easy for them. 
However, transforming an element (\ie, an image) involves editing the image, which requires proficient image editing skills. 
Therefore, we only include the adding and removing operations.

\underline{\emph{\normalsize Objective for deleting}}.
When users delete $N_d$ images in a class, 
the proxy images generated by the mutated prompt should be away from the deleted images while close to the remaining images of this class.
Moreover, according to the large margin theory~\cite{boser1992training}, images with high confidence are benefiting to the model training.
Therefore, the evaluation objective for deleting is 
\begin{equation}
    -\sum_{i}^{N_{g}}\sum_{j}^{N_{d}}\mathrm{sim}(I_{i}^{g},I_{j}^{c}) + \sum_{i}^{N_{g}}\sum_{j=N_d}^{N_{c}}\mathrm{sim}(I_{i}^{g},I_{j}^{c}) + \sum_{i}^{N_{g}}\mathrm{Confidence}(I_{i}^{g}).
\end{equation}
Here, $N_g$ is the number of proxy images, and $N_c$ represents the number of images in the class to which the deleted images belong.
$I_{i}^{g}$ and $I_{j}$ are the $i$-th proxy images and $j$-th images in the focused class, respectively.

\underline{\emph{\normalsize Objective for adding}}.
When users select $N_a$ images to add more images, the proxy images generated by the mutated prompt should have high diversity while not being away from the selected samples. 
High confidence is also encouraged. 
Therefore, the objective for adding is 
\begin{equation}
    \mathrm{S}_{div}(\mathbf{I}^g) + \sum_{i}^{N_{g}}\sum_{j}^{N_{a}}\mathrm{sim}(I_{i}^{g},I_{j}^{s})+\sum_{i}^{N_{g}}\mathrm{Confidence}(I_{i}^{g}).
\end{equation}
$\mathrm{S}_{div}$ is the diversity of the newly generated images as defined in Eq.~(\ref{eq:div}).
$\mathbf{I}^g=\{I_{1}^{g}, I_{2}^{g},...\}$ is the set of  proxy images.

\begin{table*}[t]
\centering
\caption{Comparison between projection methods in terms of both intra- and inter-modal measures (the higher, the better).}
\begin{tabular}{c|c|ccccc|ccccc}
\hline
 & \multirow{3}{*}{Methods} & \multicolumn{5}{c|}{Pets} & \multicolumn{5}{c}{COCO} \\ \cline{3-12} 
 &  & \multicolumn{2}{c|}{Intra-modal} & \multicolumn{3}{c|}{Inter-modal} & \multicolumn{2}{c|}{Intra-modal} & \multicolumn{3}{c}{Inter-modal} \\
 &  & T(30) & \multicolumn{1}{c|}{C(30)} & IMS & T(30) & C(30) & T(30) & \multicolumn{1}{c|}{C(30)} & IMS & T(30) & C(30) \\ \hline
\multirow{5}{*}{Single-modal} & CDR & 0.9633 & \multicolumn{1}{c|}{0.9758} & 0.4148 & 0.5350 & 0.6562 & 0.9593 & \multicolumn{1}{c|}{0.9640} & 0.5118 & 0.5135 & 0.5682 \\
 & t-SNE & \textbf{0.9696} & \multicolumn{1}{c|}{\textbf{0.9776}} & 0.4719 & 0.5243 & 0.6466 & \textbf{0.9708} & \multicolumn{1}{c|}{0.9675} & 0.4650 & 0.5073 & 0.5839 \\
 & UMAP & 0.9611 & \multicolumn{1}{c|}{0.9752} & 0.3874 & 0.5272 & 0.6767 & 0.9598 & \multicolumn{1}{c|}{\textbf{0.9688}} & 0.2639 & 0.4963 & 0.5571 \\
 & PCA & 0.8023 & \multicolumn{1}{c|}{0.8697} & 0.2485 & 0.5282 & 0.7175 & 0.8001 & \multicolumn{1}{c|}{0.9305} & 0.6616 & 0.5257 & 0.6358 \\
 & MDS & 0.8866 & \multicolumn{1}{c|}{0.9109} & 0.5181 & 0.5307 & 0.6821 & 0.7869 & \multicolumn{1}{c|}{0.8108} & 0.5323 & 0.5153 & 0.5264 \\ \hline
\multirow{3}{*}{Multi-modal} & DCM & 0.8804 & \multicolumn{1}{c|}{0.9011} & 0.6825 & 0.5318 & 0.6821 & 0.8037 & \multicolumn{1}{c|}{0.8326} & 0.7055 & 0.5187 & 0.5406 \\
 & MFM & 0.9132 & \multicolumn{1}{c|}{0.9537} & 0.5052 & 0.5446 & 0.7080 & 0.8154 & \multicolumn{1}{c|}{0.9040} & 0.5583 & 0.5008 & 0.6102 \\
 & Ours & 0.9627 & \multicolumn{1}{c|}{0.9740} & \textbf{0.7311} & \textbf{0.5792} & \textbf{0.7723} & 0.9569 & \multicolumn{1}{c|}{0.9663} & \textbf{0.8602} & \textbf{0.5318} & \textbf{0.7365} \\ \hline
\end{tabular}
\label{tab:quantitative}
\end{table*}

\begin{figure*}[b]
    \centering
    \includegraphics[width=\linewidth]{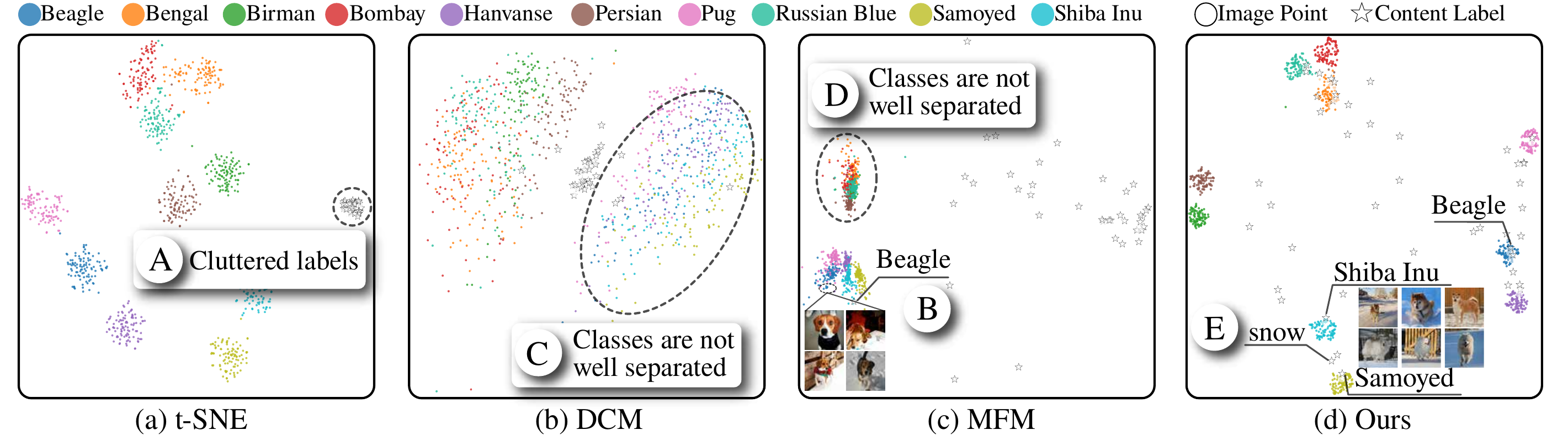}
    \caption{Visual comparison between our method and three existing methods.
    }
    \label{fig:projection}
\end{figure*}

\subsubsection{Prompt Inspection}
Users can inspect the prompts in the prompt visualization (Fig.~\ref{fig:system}(d)).
Each prompt is placed in a card.
Users can click $\vcenter{\hbox{\includegraphics[height=1.5\fontcharht\font`\B]{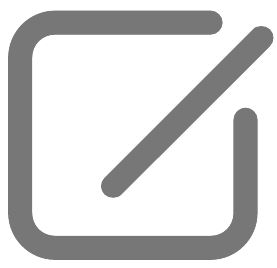}}}$ in front of the card to edit the prompt manually, or click $\vcenter{\hbox{\includegraphics[height=1.5\fontcharht\font`\B]{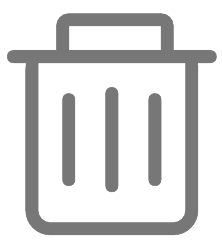}}}$ to delete it.
If a new prompt is recommended, a $\vcenter{\hbox{\includegraphics[height=1.5\fontcharht\font`\B]{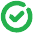}}}$ and a $\vcenter{\hbox{\includegraphics[height=1.5\fontcharht\font`\B]{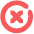}}}$ will appear for acceptance or rejection.
Users can click the rounded rectangles in the top of this visualization to filter out the prompts of a certain class.

\subsubsection{Image Filtering}
Several filters are provided to help users find images/labels of interest in the information panel (Fig.~\ref{fig:system}(a)).
Two bar charts are used to display the number distributions of original and generated images for each class.
Users can click one bar to filter out the associated original/generated images in the multi-modal distribution visualization.
The labels and their frequencies are displayed below as a list.
The experts are also interested in the ratio of generated images to original images because high ratios indicate imbalances between them.
Therefore, this list allows users to re-rank the labels based on the values of this ratio.
\section{Evaluation}
\label{evaluation}
To demonstrate the effectiveness of the proposed M2M for multi-modal projection and {\sys} for improving dataset expansion, we conducted a multi-modal data projection experiment and a case study.

\subsection{Multi-modal Projection Experiment}
\label{subsec:mmpe}

\myparagraph{Datasets}.
We conducted the experiments on two real-world datasets: Pets~\cite{parkhi2012pets} and COCO17~\cite{lin2014coco}.
The Pets dataset contains $3,842$ images from $37$ classes.
We selected ten classes of dogs and cats and $100$ images for each of the selected classes to conduct the experiment.
To obtain the content labels for each image, we first utilized GPT-4 to generate captions.
Then, we used the NLTK toolkit~\cite{loper2002nltk} to extract the nouns in the captions as the content labels.
The COCO17 dataset involves 80 classes, from which we selected all ten animal classes for the experiment.
For each selected class, we randomly sampled 1000 images.
As the images in the COCO17 dataset have captions, we directly used the NLTK toolkit~\cite{loper2002nltk} for content label extraction.

\myparagraph{Experimental settings}.
We compared the proposed M2M with two types of baselines: single-modal projection and multi-modal projection methods.
Single-modal projection methods include MDS~\cite{borg2007mds}, t-SNE~\cite{van2008tsne}, UMAP~\cite{mcinnes2018umap}, PCA~\cite{wold1987pca}, and CDR~\cite{xia2022cdr}.
To apply these methods for multi-modal projection, we utilized CLIP to extract the embeddings of images and labels in a joint space and treated both images and labels equally during projection.
Multi-modal projection methods include DCM~\cite{cheng2015dcm} and MFM~\cite{ye2024mfm}.

\begin{figure*}[b]
    \centering
    \includegraphics[width=\linewidth]{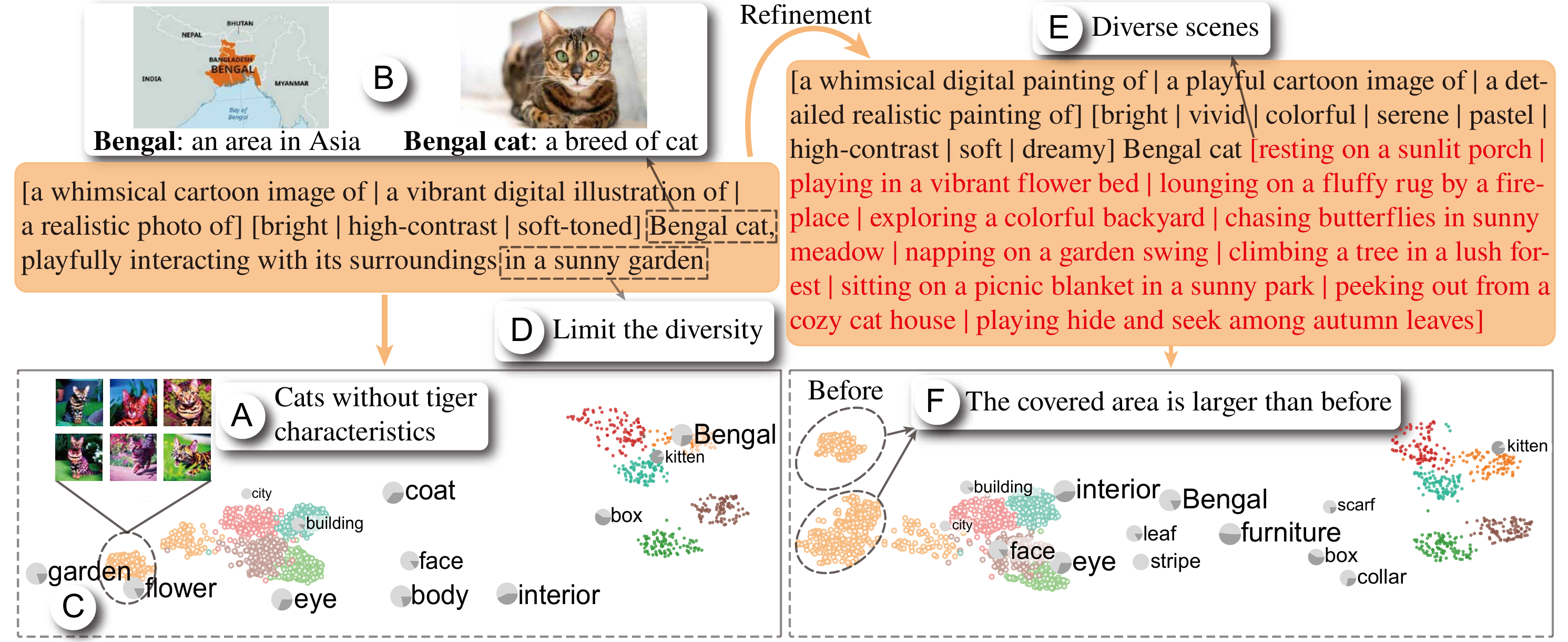}
    \caption{The prompt templates and multi-modal distribution visualization (left) before improving diversity and (right) after improving diversity.
    }
    \label{fig:comparasion}
\end{figure*}

\myparagraph{Evaluation measures}.
Similar to the work of Ye~\etal~\cite{ye2024mfm}, 
We evaluated the results using the trustworthiness and continuity metrics under both intra-modal and inter-modal settings.
To better evaluate the intra-model performance, we also propose an inter-modal similarity (IMS) measure to evaluate how well the images are placed close to their associated content labels.

\underline{\emph{\normalsize Intra- and inter-modal trustworthiness}}.
Trustworthiness measures how accurately the $k$-nearest neighbors of a point in the embedding space match the actual neighbors in the high-dimensional space.
For intra-modal trustworthiness, only the neighborhoods within images are considered.
For inter-modal trustworthiness, only the neighborhoods between images and content labels are considered.
Similar to MFM, only 30-nearest neighbors are considered (\ie, T(30)) \changjian{to balance
effectiveness and computational efficiency}.


\underline{\emph{\normalsize Intra- and inter-modal continuity}}.
Continuity measures how well the $k$-nearest neighbors of a point in the high-dimensional space are maintained in the embedding space. 
Similar to the trustworthiness, the intra- and inter-modal continuity only consider the neighborhoods within images and between images and labels, respectively.
Similar to MFM, only 30-nearest neighbors are considered (\ie, C(30)).

\underline{\emph{\normalsize Inter-modal similarity (IMS)}}.
IMS computes the average similarity between the images and their content labels in the embedding space. 

\myparagraph{Quantitative Results}.
The quantitative results are shown in Tab.~\ref{tab:quantitative}.
It can be seen that our method consistently outperforms other methods in inter-modal measures. Additionally, it demonstrates competitive intra-modal performance against single-modal projection methods, with less than a 1.5\% gap compared to the best one.
Compared with the state-of-the-art multi-modal projection method, MFM, our method performs better in all five measures.
This validates the correctness of our theoretical analysis and demonstrates the effectiveness of contrastive learning in multi-modal projection.

\myparagraph{Qualitative Results}.
We also visually compared our method with three existing methods, including one representative single-modal projection method, t-SNE, and all the two multi-modal projection methods, DCM and MFM.
The comparison is conducted on the Pets dataset.
As shown in Fig.~\ref{fig:projection}, 
t-SNE shows a clear class separation.
However, t-SNE clutters all content labels in a small area (Fig.~\ref{fig:projection}A), which demonstrates its inefficiency in preserving inter-modal relationships. 
For DCM and MFM, the inter-modal relationships are preserved to some extent.
For example, as shown in Fig.~\ref{fig:projection}B, the content label ``Beagle'' is placed close to the images of beagles.
However, some of the classes are not well separated (\eg, Figs.~\ref{fig:projection}C and ~\ref{fig:projection}D).
This is because the multi-modal distance order loss cannot be well optimized in this many-to-many setting, which inversely affects the class separation optimization.
Our method not only preserves intra-modal relationships (\ie, a clear class separation in Fig.~\ref{fig:projection}E), but also inter-modal relationships in the many-to-many setting.
For example, the Shiba Inu and Samoyed dogs tend to appear in snow settings.
Thus, the content label ``snow'' is placed in the middle of classes ``Shiba Inu'' and ``Samoyed''.


\subsection{Case Study}
We conducted a case study with E1, one of the experts in the requirement analysis, to demonstrate the effectiveness of {\sys} in human-guided image generation for better training image datasets.
The case study was conducted on the ten classes of dogs and cats \feilvvl{of the Pets} dataset used in the multi-modal projection experiment (Sec.~\ref{subsec:mmpe}).
The ten classes included Beagle, Bengal, Birman, Bombay, Havanese, Persian, Pug, Russian Blue, Samoyed, and Shiba Inu. 
Each class contained 100 images. 
Initially, we trained a classifier on the $10\times 100$ images, which achieved an accuracy of 48.45\%.
To improve accuracy, E1 chose to use our tool to expand the training data. 
Since E1 was not involved in the design phase, we provided him with an overview of the visual design and interactions of {\sys} before the case study.
In the case study, we used the pair analytics protocol~\cite{arias2011pair}, which enables experts to focus on analytical tasks while we handle the navigation of the tool.

\myparagraph{Overview}.
Initially, to understand how the generated images looked like with the default prompt templates, E1 generated 200 images for each class.
These generated images, together with the original images, were projected with their content labels by our M2M method.
As shown in Fig.~\ref{fig:system}(d), the images are divided into two main sections: the upper and lower parts.
According to the content labels, E1 quickly knew that the images in the upper part were mainly cats (\eg, ``Bengal'' and ``kitten'' in Fig.~\ref{fig:system}A), and the images in the lower part were mainly dogs (\eg, ``Shiba Inu'' and ``Samoyed'' in Fig.~\ref{fig:system}B).
This result met E1's expectation, as the distinction between dogs and cats is greater than the differences within each class.

During the examination of the content labels, E1 found that in the pie charts of some content labels (\eg $\vcenter{\hbox{\includegraphics[height=1.5\fontcharht\font`\B]{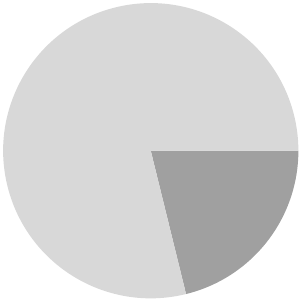}}}$ body, and $\vcenter{\hbox{\includegraphics[height=1.5\fontcharht\font`\B]{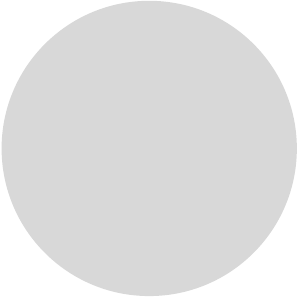}}}$ stripe), the light gray sector occupied much more than 2/3.
It was quite abnormal because the generated images were twice as many as the original ones. 
To find more such content labels, E1 re-ranked the content labels in Fig.~\ref{fig:system}C by the number \feilvvl{of the generated} images over the number of original images ratio. 
The content label ``tiger'' was ranked first, with the light gray sector occupying all of the pie chart (Fig.~\ref{fig:system}D).
It indicated that tigers only appeared in the generated images. 
To understand the underlying causes, E1 selected this label and checked the associated images.

\myparagraph{Identifying undesired images in class ``Bengal''}.
According to the number distribution \feilvvl{of the generated} images, E1 found that most of the images of the content label ``tiger'' belong to the class ``Bengal'' (Fig.~\ref{fig:system}E). 
Wondering why ``tiger'' appeared in this class, E1 checked these images in the detail panel (Fig.~\ref{fig:system}(e)) and found that these images had the bodies of cats but the faces of tigers (Fig.~\ref{fig:system}F).
Upon this finding, E1 first suspected that the prompts wrongly included tiger-related descriptions.
Therefore, he turned to prompt visualization to check the prompt template for the class ``Bengal.''
However, he found the prompt template was correct and had no description of tigers (Fig.~\ref{fig:system}G).
Having no idea why such cat-tiger-hybrid images were generated, E1 decided to use the sample-level prompt refinement method to recommend a new prompt template.
Thus, he selected these images and deleted them by clicking $\vcenter{\hbox{\includegraphics[height=1.5\fontcharht\font`\B]{figures/icons/delete.pdf}}}$. 
Then, a new prompt template was recommended, and 200 more images were generated for the class ``Bengal'' with this new prompt template.
E1 checked the newly generated images and found they were cats without tiger characteristics (Fig.~\ref{fig:comparasion}A).

Wondering why the new prompt template could correctly generate cats, 
E1 checked the new prompt template and found an interesting modification.
The name of the class was changed from ``Bengal'' to ``Bengal cat'' (Fig.~\ref{fig:comparasion}B).
It reminded E1 that Bengal is the name of an area in Asia instead of a breed of cat.
One of the most famous breeds in this area is the Bengal tiger.
Because the data class name did not explicitly specify the Bengal cats, the model might wrongly interpret it as Bengal tigers.
This led to the generated images featuring tiger characteristics. 
E1 felt that the change to the prompt template was correct and accepted it. 
He also checked the class names of the other nine classes and found one class, ``Bombay'' (a city in India), with a similar issue. 
For this class, he changed the class name to ``Bombay cat'' and regenerated the images.

\myparagraph{Improving the diversity of class ``Bengal''}.
After that, E1 found in the metric visualization that the informativeness metric increased, and the distance metric decreased, which met his expectations (Fig.~\ref{fig:system}H1).
However, the diversity metric decreased, which might have limited the model's performance~\cite{shi2016diversifying}.
To understand why the diversity metric decreased, E1 first checked the newly generated images of class ``Bengal'' and found new content labels ``garden'' and ``flower'' arose in this area (Fig.~\ref{fig:comparasion}C).
It \feilvvl{reminded} him that the prompt template contained ``in a sunny garden'' (Fig.~\ref{fig:comparasion}D), which limited the scene of the generated images, thus decreasing the diversity.
Based on this finding, E1 decided to increase the diversity of these generated images.
Therefore, he selected these newly generated images of class ``Bengal'' and clicked $\vcenter{\hbox{\includegraphics[height=1.5\fontcharht\font`\B]{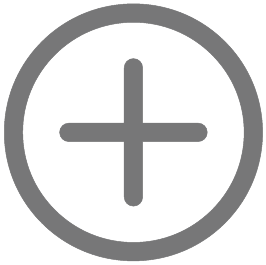}}}$ to add more diverse generated images.
Then, a new prompt template was recommended, and 200 more images were generated for the class ``Bengal'' with this new prompt template.
The scene descriptions in the new template had become more diverse and were not limited to just including gardens (Fig.~\ref{fig:comparasion}E).
After that, the diversity increased  (Fig.~\ref{fig:system}H2).
Moreover, in the multi-modal distribution visualization, the generated images covered a larger area (Fig.~\ref{fig:comparasion}F).
Both of these patterns indicated an increase in diversity.
Therefore, E1 thought this prompt template was useful and accepted it.

Similar to the analysis of class ``Bengal,'' E1 analyzed the remaining nine classes and modified the corresponding prompts with the sample-level prompt refinement method.

\subsection{Post Analysis}

\myparagraph{Effectiveness in improving classification performance}.
After the case study, we conducted a quantitative evaluation to demonstrate the effectiveness of {\sys}. 
We trained a classifier (ResNet50~\cite{he2016deep}) on the generated images after each step in the case study and tested its accuracy on the test set of the Pets dataset.
The results are shown in Table~\ref{tab:case}.
It can be seen that the accuracy of the classifier was steadily improved after each step, demonstrating the robustness of the prompts generated by our method. 
In summary, using our method improves the accuracy performance from 48.45\% to 81.80\%, showcasing the effectiveness of our approach.

\myparagraph{Comparison with automatic dataset expansion method}.
To better demonstrate the advantages of our human-guided dataset expansion method compared to existing automatic methods, we conducted a comparative experiment.
In this comparison, our method was evaluated against GIF~\cite{zhang2023expanding}, which is one of the leading dataset expansion methods. 
For the GIF method, we used the default prompts to generate ten times the number of original images.
For our method, to ensure a fair comparison, we used the prompts modified in the case study to generate the same number of images. 
The results are shown in the last two lines of Table~\ref{tab:case}.
Our method improved accuracy by approximately 5\% over GIF. 
This improvement is attributed to our method's ability to quickly identify issues with the images generated by the current prompts (\eg, undesired tiger-liked cat images), which allows for timely adjustments for better dataset expansion.


\myparagraph{Extension to other tasks}.
The dataset expanded by our method can also be utilized for tasks beyond classification, such as object detection~\cite{liu2016towards, chen2022towards, zhang2024adapt}. 
To illustrate this, we used the generated images in the case study to enhance the performance of an object detector.
Initially, we trained an SSD detection model~\cite{liu2016ssd} using the original images, achieving a mean Average Precision (mAP) of 92.4\%. This high mAP is partly due to the use of a backbone pre-trained on ImageNet, which is a common practice in object detection.
Next, we combined the original images with the generated images to retrain the SSD detection model. Since the generated images did not have bounding box annotations, we employed the commonly used pseudo-labeling method for training. Specifically, we used the detector trained on the original images to generate bounding box annotations for the generated images. We then treated the high-confidence generated bounding box annotations as ground truth and added them to the training dataset.
As a result of incorporating the generated images, the mAP improved to 94.5\%. 
Given that the original mAP was already high, this improvement is quite significant, demonstrating the generalization ability of our method to other tasks.

\begin{table}[t]
\caption{Performance Improvement With {\sys}.}
\label{tab:case}
\begin{tabularx}{\linewidth}{>{\raggedright\arraybackslash}p{3.8cm} >{\centering\arraybackslash}p{2.5cm} >{\centering\arraybackslash}p{1.2cm}} 
\toprule
Step & \# Generated images & Accuracy \\ \midrule
Base & 0 & 48.45\% \\ \hline
Identifying undesired images in class “Bengal” & \vspace*{\fill}2,065\vspace*{\fill} & \vspace*{\fill}51.00\%\vspace*{\fill} \\ \hline
Improving the diversity of class “Bengal” & \vspace*{\fill}2,265\vspace*{\fill} & \vspace*{\fill}52.97\%\vspace*{\fill} \\ \bottomrule \hline
Dataset expansion without human guidance & \vspace*{\fill}30,000\vspace*{\fill} & \vspace*{\fill}76.03\%\vspace*{\fill} \\ \hline
Dataset expansion with human guidance (our method) & \vspace*{\fill}30,000\vspace*{\fill} & \vspace*{\fill}\textbf{81.80\%}\vspace*{\fill} \\ \bottomrule
\end{tabularx}
\end{table}
\section{Expert Feedback and Discussion}
\label{sec:discuss}
Following the case study, we conducted four interviews to gather feedback from a group of experts. This group included the two experts (E1, E2) who collaborated with us in the requirement analysis and two new experts we invited (E5, E6).
The newly invited experts were Ph.D. students who had over two years of experience in computer vision research.
Each interview lasted between 40 and 65 minutes. 
Overall, the expert feedback was positive regarding the usability of {\sys}, but some limitations were also identified, pointing out areas that require further investigation in the future.

\subsection{Usability}

\myparagraph{Efficient exploration of images}.
E6 appreciated the efficiency gained through the projection-based exploration method, commenting, ``This tool greatly reduces the time and effort needed to explore large datasets of generated images. By projecting images and their content labels in a 2D space, I can quickly spot the outliers or less relevant images, making the identification of undesired samples much easier.''
E5 also highlighted the convenience of the content labels. ``Instead of manually reviewing each image, I can focus on images of interest and make more informed refinements to the prompts.'' He said.


\myparagraph{Easy prompt refinement}.
E1 appreciated how DataCrafter simplified the prompt refinement process. 
He said, ``Instead of struggling to adjust the text prompts, I can now give simple feedback on individual images. 
This method lowers the barrier to generate diverse and relevant images for the dataset." 
E5 also found the process efficient, commenting, “Combined with the multi-modal distribution visualization, the sample-level prompt refinement not only saves time but also ensures that the generated images match my requirements more closely.''


\subsection{Limitations and Future Work}


\myparagraph{Operations on multiple sets}.
The current sample-level prompt refinement method primarily focuses on providing feedback for a single set of generated images. 
According to set theory, there are many operations on multiple sets, which are more complex and provide richer feedback opportunities.
For example, we can combine two sets of images to generate more images that fuse the styles of the two sets.
Exploring how feedback can be effectively applied across multiple sets could enhance the diversity and quality of the generated images, 
which is an interesting venue for future work.


\myparagraph{\changjian{Generalization to other types of images}}.
\changjian{In the evaluation, we demonstrated the effectiveness of our method in improving the quality of natural image datasets.
Additionally, it can also be used for other types of images as long as the corresponding generative models are available.
However, because generative models for other image types (\eg, CXRL~\cite{han2024advancing} for medical images) are generally less developed than those for natural images, the quality of generated images for these types tends to be more problematic.
Therefore, it would be valuable for future research to explore how these more severe data quality issues impact the effectiveness of {\sys} and investigate ways to improve it.}

\myparagraph{\changjian{Evaluation on efficiency}}.
\changjian{During the case study, we focused on assessing the effectiveness of the proposed {\sys} in improving the quality of generated images. 
However, efficiency also plays a crucial role in practical applications. 
If the tool takes too long to use, even with improved performance, users may be reluctant to adopt it. 
Therefore, it would be valuable to conduct user studies to explore the interplay between effectiveness and efficiency, and to identify an optimal balance between the two.
}

\section{Conclusion}

In conclusion, this paper presents a novel human-guided image generation method to address the limitations of automatic image generation for dataset expansion in computer vision tasks. 
By leveraging a multi-modal projection method with theoretical guarantees, users are able to efficiently explore both original and generated images, which helps find quality issues in the generated images. 
Additionally, the introduction of a sample-level prompt refinement method simplifies the process of adjusting prompts, making it more accessible for users to enhance the quality of generated images. 
Several quantitative experiments and a case study were conducted to demonstrate the effectiveness of the proposed method.


\acknowledgments{
The work is supported by the National Natural Science Foundation of China (Grant Nos. 62225205, 62402167, and 92055213), the Science and Technology Innovation Program of Hunan Province (No. 2023ZJ1080), the Science and Technology Program of Changsha (kh2301011), the Shenzhen Basic Research Project (Natural Science Foundation) (JCYJ20210324140002006), and the Innovative Development Project of Hunan Meteorological Bureau (CXFZ2024-FZZX27).
}

\bibliographystyle{abbrv-doi-hyperref}

\bibliography{reference}

\begin{thebibliography}{10}

\bibitem{arias2011pair}
R.~Arias-Hernandez, L.~T. Kaastra, T.~M. Green, and B.~Fisher.
\newblock Pair analytics: Capturing reasoning processes in collaborative visual analytics.
\newblock In {\em IEEE Hawaii International Conference on System Sciences}, pp. 1--10, 2011. \href{https://doi.org/10.1109/HICSS.2011.339}
{doi: {{%
10\hspace{.1pt}\discretionary{.}{%
}{.}\hspace{.4pt}1109\discretionary{/}{%
}{/}HICSS\hspace{.1pt}\discretionary{.}{%
}{.}\hspace{.4pt}2011\hspace{.1pt}\discretionary{.}{%
}{.}\hspace{.4pt}339}}}


\bibitem{borg2007mds}
I.~Borg and P.~J. Groenen.
\newblock {\em Modern multidimensional scaling: Theory and applications}.
\newblock Springer Science \& Business Media, 2007. \href{https://doi.org/10.1111/j.1745-3984.2003.tb01108.x}
{doi: {{%
10\hspace{.1pt}\discretionary{.}{%
}{.}\hspace{.4pt}1111\discretionary{/}{%
}{/}j\hspace{.1pt}\discretionary{.}{%
}{.}\hspace{.4pt}1745\discretionary{%
}{-}{-}3984\hspace{.1pt}\discretionary{.}{%
}{.}\hspace{.4pt}2003\hspace{.1pt}\discretionary{.}{%
}{.}\hspace{.4pt}tb01108\hspace{.1pt}\discretionary{.}{%
}{.}\hspace{.4pt}x}}}


\bibitem{boser1992training}
B.~E. Boser, I.~M. Guyon, and V.~N. Vapnik.
\newblock A training algorithm for optimal margin classifiers.
\newblock In {\em Proceedings of the Workshop on Computational Learning Theory}, pp. 144--152, 1992. \href{https://doi.org/10.1145/130385.130401}
{doi: {{%
10\hspace{.1pt}\discretionary{.}{%
}{.}\hspace{.4pt}1145\discretionary{/}{%
}{/}130385\hspace{.1pt}\discretionary{.}{%
}{.}\hspace{.4pt}130401}}}


\bibitem{brade2023promptify}
S.~Brade, B.~Wang, M.~Sousa, S.~Oore, and T.~Grossman.
\newblock {Promptify}: Text-to-image generation through interactive prompt exploration with large language models.
\newblock In {\em Proceedings of the Annual ACM Symposium on User Interface Software and Technology}, pp. 1--14, 2023. \href{https://doi.org/10.1145/3586183.3606725}
{doi: {{%
10\hspace{.1pt}\discretionary{.}{%
}{.}\hspace{.4pt}1145\discretionary{/}{%
}{/}3586183\hspace{.1pt}\discretionary{.}{%
}{.}\hspace{.4pt}3606725}}}


\bibitem{chen2024enhancing}
C.~Chen, J.~Chen, W.~Yang, H.~Wang, J.~Knittel, X.~Zhao, S.~Koch, T.~Ertl, and S.~Liu.
\newblock Enhancing single-frame supervision for better temporal action localization.
\newblock {\em IEEE Transactions on Visualization and Computer Graphics}, 30(6):2903--2915, 2024. \href{https://doi.org/10.1109/TVCG.2024.3388521}
{doi: {{%
10\hspace{.1pt}\discretionary{.}{%
}{.}\hspace{.4pt}1109\discretionary{/}{%
}{/}TVCG\hspace{.1pt}\discretionary{.}{%
}{.}\hspace{.4pt}2024\hspace{.1pt}\discretionary{.}{%
}{.}\hspace{.4pt}3388521}}}


\bibitem{chen2024unified}
C.~Chen, Y.~Guo, F.~Tian, S.~Liu, W.~Yang, Z.~Wang, J.~Wu, H.~Su, H.~Pfister, and S.~Liu.
\newblock A unified interactive model evaluation for classification, object detection, and instance segmentation in computer vision.
\newblock {\em IEEE Transactions on Visualization and Computer Graphics}, 30(1):76--86, 2024. \href{https://doi.org/10.1109/TVCG.2023.3326588}
{doi: {{%
10\hspace{.1pt}\discretionary{.}{%
}{.}\hspace{.4pt}1109\discretionary{/}{%
}{/}TVCG\hspace{.1pt}\discretionary{.}{%
}{.}\hspace{.4pt}2023\hspace{.1pt}\discretionary{.}{%
}{.}\hspace{.4pt}3326588}}}


\bibitem{chen2021interactive}
C.~Chen, Z.~Wang, J.~Wu, X.~Wang, L.-Z. Guo, Y.-F. Li, and S.~Liu.
\newblock Interactive graph construction for graph-based semi-supervised learning.
\newblock {\em IEEE Transactions on Visualization and Computer Graphics}, 27(9):3701--3716, 2021. \href{https://doi.org/10.1109/TVCG.2021.3084694}
{doi: {{%
10\hspace{.1pt}\discretionary{.}{%
}{.}\hspace{.4pt}1109\discretionary{/}{%
}{/}TVCG\hspace{.1pt}\discretionary{.}{%
}{.}\hspace{.4pt}2021\hspace{.1pt}\discretionary{.}{%
}{.}\hspace{.4pt}3084694}}}


\bibitem{chen2022towards}
C.~Chen, J.~Wu, X.~Wang, S.~Xiang, S.-H. Zhang, Q.~Tang, and S.~Liu.
\newblock Towards better caption supervision for object detection.
\newblock {\em IEEE Transactions on Visualization and Computer Graphics}, 28(4):1941--1954, 2022. \href{https://doi.org/10.1109/TVCG.2021.3138933}
{doi: {{%
10\hspace{.1pt}\discretionary{.}{%
}{.}\hspace{.4pt}1109\discretionary{/}{%
}{/}TVCG\hspace{.1pt}\discretionary{.}{%
}{.}\hspace{.4pt}2021\hspace{.1pt}\discretionary{.}{%
}{.}\hspace{.4pt}3138933}}}


\bibitem{chen2020oodanalyzer}
C.~Chen, J.~Yuan, Y.~Lu, Y.~Liu, H.~Su, S.~Yuan, and S.~Liu.
\newblock {OoDAnalyzer}: Interactive analysis of out-of-distribution samples.
\newblock {\em IEEE Transactions on Visualization and Computer Graphics}, 27(7):3335--3349, 2021. \href{https://doi.org/10.1109/TVCG.2020.2973258}
{doi: {{%
10\hspace{.1pt}\discretionary{.}{%
}{.}\hspace{.4pt}1109\discretionary{/}{%
}{/}TVCG\hspace{.1pt}\discretionary{.}{%
}{.}\hspace{.4pt}2020\hspace{.1pt}\discretionary{.}{%
}{.}\hspace{.4pt}2973258}}}


\bibitem{cheng2015dcm}
S.~Cheng and K.~Mueller.
\newblock The data context map: Fusing data and attributes into a unified display.
\newblock {\em IEEE Transactions on Visualization and Computer Graphics}, 22(1):121--130, 2015. \href{https://doi.org/10.1109/tvcg.2015.2467552}
{doi: {{%
10\hspace{.1pt}\discretionary{.}{%
}{.}\hspace{.4pt}1109\discretionary{/}{%
}{/}tvcg\hspace{.1pt}\discretionary{.}{%
}{.}\hspace{.4pt}2015\hspace{.1pt}\discretionary{.}{%
}{.}\hspace{.4pt}2467552}}}


\bibitem{chung2023promptpaint}
J.~J.~Y. Chung and E.~Adar.
\newblock {PromptPaint}: Steering text-to-image generation through paint medium-like interactions.
\newblock In {\em Proceedings of the Annual ACM Symposium on User Interface Software and Technology}, pp. 23--32, 2023. \href{https://doi.org/10.1145/3586183.3606777}
{doi: {{%
10\hspace{.1pt}\discretionary{.}{%
}{.}\hspace{.4pt}1145\discretionary{/}{%
}{/}3586183\hspace{.1pt}\discretionary{.}{%
}{.}\hspace{.4pt}3606777}}}


\bibitem{dunlap2023diversify}
L.~Dunlap, A.~Umino, H.~Zhang, J.~Yang, J.~E. Gonzalez, and T.~Darrell.
\newblock Diversify your vision datasets with automatic diffusion-based augmentation.
\newblock In {\em Proceedings of the Advances in Neural Information Processing Systems}, pp. 79024--79034, 2023.

\bibitem{endert2011observation}
A.~Endert, C.~Han, D.~Maiti, L.~House, and C.~North.
\newblock Observation-level interaction with statistical models for visual analytics.
\newblock In {\em Proceedings of the IEEE Conference on Visual Analytics Science and Technology}, pp. 121--130, 2011. \href{https://doi.org/10.1109/vast.2011.6102449}
{doi: {{%
10\hspace{.1pt}\discretionary{.}{%
}{.}\hspace{.4pt}1109\discretionary{/}{%
}{/}vast\hspace{.1pt}\discretionary{.}{%
}{.}\hspace{.4pt}2011\hspace{.1pt}\discretionary{.}{%
}{.}\hspace{.4pt}6102449}}}


\bibitem{feng2024promptmagician}
Y.~Feng, X.~Wang, K.~K. Wong, S.~Wang, Y.~Lu, M.~Zhu, B.~Wang, and W.~Chen.
\newblock {PromptMagician}: Interactive prompt engineering for text-to-image creation.
\newblock {\em IEEE Transactions on Visualization and Computer Graphics}, 30(1):295--305, 2024. \href{https://doi.org/10.1109/tvcg.2023.3327168}
{doi: {{%
10\hspace{.1pt}\discretionary{.}{%
}{.}\hspace{.4pt}1109\discretionary{/}{%
}{/}tvcg\hspace{.1pt}\discretionary{.}{%
}{.}\hspace{.4pt}2023\hspace{.1pt}\discretionary{.}{%
}{.}\hspace{.4pt}3327168}}}


\bibitem{furnas1986generalized}
G.~W. Furnas.
\newblock Generalized fisheye views.
\newblock {\em ACM Sigchi Bulletin}, 17(4):16--23, 1986.

\bibitem{gou2020vatld}
L.~Gou, L.~Zou, N.~Li, M.~Hofmann, A.~K. Shekar, A.~Wendt, and L.~Ren.
\newblock {VATLD}: A visual analytics system to assess, understand and improve traffic light detection.
\newblock {\em IEEE Transactions on Visualization and Computer Graphics}, 27(2):261--271, 2021. \href{https://doi.org/10.1109/TVCG.2020.3030350}
{doi: {{%
10\hspace{.1pt}\discretionary{.}{%
}{.}\hspace{.4pt}1109\discretionary{/}{%
}{/}TVCG\hspace{.1pt}\discretionary{.}{%
}{.}\hspace{.4pt}2020\hspace{.1pt}\discretionary{.}{%
}{.}\hspace{.4pt}3030350}}}


\bibitem{guoconnecting}
Q.~Guo, R.~Wang, J.~Guo, B.~Li, K.~Song, X.~Tan, G.~Liu, J.~Bian, and Y.~Yang.
\newblock Connecting large language models with evolutionary algorithms yields powerful prompt optimizers.
\newblock In {\em The Proceedings of the International Conference on Learning Representations}, 2024.

\bibitem{guo2024prompTHis}
Y.~Guo, H.~Shao, C.~Liu, K.~Xu, and X.~Yuan.
\newblock {PrompTHis}: Visualizing the process and influence of prompt editing during text-to-image creation.
\newblock {\em IEEE Transactions on Visualization and Computer Graphics}, 2024.
\newblock to be published. \href{https://doi.org/10.1109/tvcg.2024.3408255}
{doi: {{%
10\hspace{.1pt}\discretionary{.}{%
}{.}\hspace{.4pt}1109\discretionary{/}{%
}{/}tvcg\hspace{.1pt}\discretionary{.}{%
}{.}\hspace{.4pt}2024\hspace{.1pt}\discretionary{.}{%
}{.}\hspace{.4pt}3408255}}}


\bibitem{han2024advancing}
W.~Han, C.~Kim, D.~Ju, Y.~Shim, and S.~J. Hwang.
\newblock Advancing text-driven chest x-ray generation with policy-based reinforcement learning.
\newblock In {\em Proceedings of the International Conference on Medical Image Computing and Computer-Assisted Intervention}, pp. 56--66, 2024.

\bibitem{hasani2022artificial}
N.~Hasani, F.~Farhadi, M.~A. Morris, M.~Nikpanah, A.~Rahmim, Y.~Xu, A.~Pariser, M.~T. Collins, R.~M. Summers, E.~Jones, et~al.
\newblock Artificial intelligence in medical imaging and its impact on the rare disease community: threats, challenges and opportunities.
\newblock {\em PET Clinics}, 17(1):13--29, 2022. \href{https://doi.org/10.1016/j.cpet.2021.09.009}
{doi: {{%
10\hspace{.1pt}\discretionary{.}{%
}{.}\hspace{.4pt}1016\discretionary{/}{%
}{/}j\hspace{.1pt}\discretionary{.}{%
}{.}\hspace{.4pt}cpet\hspace{.1pt}\discretionary{.}{%
}{.}\hspace{.4pt}2021\hspace{.1pt}\discretionary{.}{%
}{.}\hspace{.4pt}09\hspace{.1pt}\discretionary{.}{%
}{.}\hspace{.4pt}009}}}


\bibitem{hausdorff2007set}
F.~Hausdorff.
\newblock {\em Set Theory}.
\newblock Chelsea, 2007.

\bibitem{he2024videopro}
J.~He, X.~Wang, K.~K. Wong, X.~Huang, C.~Chen, Z.~Chen, F.~Wang, M.~Zhu, and H.~Qu.
\newblock {VideoPro}: A visual analytics approach for interactive video programming.
\newblock {\em IEEE Transactions on Visualization and Computer Graphics}, 30(1):87--97, 2024. \href{https://doi.org/10.1109/tvcg.2023.3326586}
{doi: {{%
10\hspace{.1pt}\discretionary{.}{%
}{.}\hspace{.4pt}1109\discretionary{/}{%
}{/}tvcg\hspace{.1pt}\discretionary{.}{%
}{.}\hspace{.4pt}2023\hspace{.1pt}\discretionary{.}{%
}{.}\hspace{.4pt}3326586}}}


\bibitem{he2016deep}
K.~He, X.~Zhang, S.~Ren, and J.~Sun.
\newblock Deep residual learning for image recognition.
\newblock In {\em IEEE Conference on Computer Vision and Pattern Recognition}, pp. 770--778, 2016. \href{https://doi.org/10.1109/CVPR.2016.90}
{doi: {{%
10\hspace{.1pt}\discretionary{.}{%
}{.}\hspace{.4pt}1109\discretionary{/}{%
}{/}CVPR\hspace{.1pt}\discretionary{.}{%
}{.}\hspace{.4pt}2016\hspace{.1pt}\discretionary{.}{%
}{.}\hspace{.4pt}90}}}


\bibitem{he2021can}
W.~He, L.~Zou, A.~K. Shekar, L.~Gou, and L.~Ren.
\newblock Where can we help? a visual analytics approach to diagnosing and improving semantic segmentation of movable objects.
\newblock {\em IEEE Transactions on Visualization and Computer Graphics}, 28(1):1040--1050, 2021. \href{https://doi.org/10.1109/TVCG.2021.3114855}
{doi: {{%
10\hspace{.1pt}\discretionary{.}{%
}{.}\hspace{.4pt}1109\discretionary{/}{%
}{/}TVCG\hspace{.1pt}\discretionary{.}{%
}{.}\hspace{.4pt}2021\hspace{.1pt}\discretionary{.}{%
}{.}\hspace{.4pt}3114855}}}


\bibitem{heusel2017gans}
M.~Heusel, H.~Ramsauer, T.~Unterthiner, B.~Nessler, and S.~Hochreiter.
\newblock {GANs} trained by a two time-scale update rule converge to a local nash equilibrium.
\newblock In {\em Proceedings of the 31st International Conference on Neural Information Processing Systems}, p. 6629–6640, 2017. \href{https://doi.org/10.18034/ajase.v8i1.9}
{doi: {{%
10\hspace{.1pt}\discretionary{.}{%
}{.}\hspace{.4pt}18034\discretionary{/}{%
}{/}ajase\hspace{.1pt}\discretionary{.}{%
}{.}\hspace{.4pt}v8i1\hspace{.1pt}\discretionary{.}{%
}{.}\hspace{.4pt}9}}}


\bibitem{hoferlin2012inter}
B.~H{\"o}ferlin, R.~Netzel, M.~H{\"o}ferlin, D.~Weiskopf, and G.~Heidemann.
\newblock Inter-active learning of ad-hoc classifiers for video visual analytics.
\newblock In {\em Proceedings of the Conference on Visual Analytics Science and Technology}, pp. 23--32, 2012. \href{https://doi.org/10.1109/vast.2012.6400492}
{doi: {{%
10\hspace{.1pt}\discretionary{.}{%
}{.}\hspace{.4pt}1109\discretionary{/}{%
}{/}vast\hspace{.1pt}\discretionary{.}{%
}{.}\hspace{.4pt}2012\hspace{.1pt}\discretionary{.}{%
}{.}\hspace{.4pt}6400492}}}


\bibitem{hoque2023visual}
M.~N. Hoque, W.~He, A.~K. Shekar, L.~Gou, and L.~Ren.
\newblock Visual concept programming: A visual analytics approach to injecting human intelligence at scale.
\newblock {\em IEEE Transactions on Visualization and Computer Graphics}, 29(1):74--83, 2023. \href{https://doi.org/10.1109/TVCG.2022.3209466}
{doi: {{%
10\hspace{.1pt}\discretionary{.}{%
}{.}\hspace{.4pt}1109\discretionary{/}{%
}{/}TVCG\hspace{.1pt}\discretionary{.}{%
}{.}\hspace{.4pt}2022\hspace{.1pt}\discretionary{.}{%
}{.}\hspace{.4pt}3209466}}}


\bibitem{jayasumana2024rethinking}
S.~Jayasumana, S.~Ramalingam, A.~Veit, D.~Glasner, A.~Chakrabarti, and S.~Kumar.
\newblock Rethinking fid: Towards a better evaluation metric for image generation.
\newblock In {\em Proceedings of the IEEE/CVF Conference on Computer Vision and Pattern Recognition (CVPR)}, pp. 9307--9315, June 2024.

\bibitem{jia2022towards}
S.~Jia, Z.~Li, N.~Chen, and J.~Zhang.
\newblock Towards visual explainable active learning for zero-shot classification.
\newblock {\em IEEE Transactions on Visualization and Computer Graphics}, 28(1):791--801, 2022. \href{https://doi.org/10.1109/TVCG.2021.3114793}
{doi: {{%
10\hspace{.1pt}\discretionary{.}{%
}{.}\hspace{.4pt}1109\discretionary{/}{%
}{/}TVCG\hspace{.1pt}\discretionary{.}{%
}{.}\hspace{.4pt}2021\hspace{.1pt}\discretionary{.}{%
}{.}\hspace{.4pt}3114793}}}


\bibitem{li2024semantic}
B.~Li, X.~Xu, X.~Wang, Y.~Hou, Y.~Feng, F.~Wang, X.~Zhang, Q.~Zhu, and W.~Che.
\newblock Semantic-guided generative image augmentation method with diffusion models for image classification.
\newblock In {\em Proceedings of the AAAI Conference on Artificial Intelligence}, pp. 3018--3027, 2024. \href{https://doi.org/10.1609/aaai.v38i4.28084}
{doi: {{%
10\hspace{.1pt}\discretionary{.}{%
}{.}\hspace{.4pt}1609\discretionary{/}{%
}{/}aaai\hspace{.1pt}\discretionary{.}{%
}{.}\hspace{.4pt}v38i4\hspace{.1pt}\discretionary{.}{%
}{.}\hspace{.4pt}28084}}}


\bibitem{li2024evovis}
S.~Li, G.~Liu, T.~Wei, S.~Jia, and J.~Zhang.
\newblock {EvoVis}: A visual analytics method to understand the labeling iterations in data programming.
\newblock {\em IEEE Transactions on Visualization and Computer Graphics}, 2024.
\newblock to be published. \href{https://doi.org/10.1109/tvcg.2024.3370654}
{doi: {{%
10\hspace{.1pt}\discretionary{.}{%
}{.}\hspace{.4pt}1109\discretionary{/}{%
}{/}tvcg\hspace{.1pt}\discretionary{.}{%
}{.}\hspace{.4pt}2024\hspace{.1pt}\discretionary{.}{%
}{.}\hspace{.4pt}3370654}}}


\bibitem{li2024visual}
Y.~Li, J.~Wang, P.~Aboagye, C.-C.~M. Yeh, Y.~Zheng, L.~Wang, W.~Zhang, and K.-L. Ma.
\newblock Visual analytics for efficient image exploration and user-guided image captioning.
\newblock {\em IEEE Transactions on Visualization and Computer Graphics}, 2024.
\newblock to be published. \href{https://doi.org/10.1109/tvcg.2024.3388514}
{doi: {{%
10\hspace{.1pt}\discretionary{.}{%
}{.}\hspace{.4pt}1109\discretionary{/}{%
}{/}tvcg\hspace{.1pt}\discretionary{.}{%
}{.}\hspace{.4pt}2024\hspace{.1pt}\discretionary{.}{%
}{.}\hspace{.4pt}3388514}}}


\bibitem{lin2014coco}
T.-Y. Lin, M.~Maire, S.~Belongie, J.~Hays, P.~Perona, D.~Ramanan, P.~Doll{\'a}r, and C.~L. Zitnick.
\newblock Microsoft coco: Common objects in context.
\newblock In {\em Proceedings of the European Conference on Computer Vision}, pp. 740--755, 2014. \href{https://doi.org/10.1007/978-3-319-10602-1_48}
{doi: {{%
10\hspace{.1pt}\discretionary{.}{%
}{.}\hspace{.4pt}1007\discretionary{/}{%
}{/}978\discretionary{%
}{-}{-}3\discretionary{%
}{-}{-}319\discretionary{%
}{-}{-}10602\discretionary{%
}{-}{-}1\_48}}}


\bibitem{liu2022memory}
J.~Liu, W.~Li, and Y.~Sun.
\newblock Memory-based jitter: Improving visual recognition on long-tailed data with diversity in memory.
\newblock {\em Proceedings of the AAAI Conference on Artificial Intelligence}, 36(2):1720–1728, 2022. \href{https://doi.org/10.1609/aaai.v36i2.20064}
{doi: {{%
10\hspace{.1pt}\discretionary{.}{%
}{.}\hspace{.4pt}1609\discretionary{/}{%
}{/}aaai\hspace{.1pt}\discretionary{.}{%
}{.}\hspace{.4pt}v36i2\hspace{.1pt}\discretionary{.}{%
}{.}\hspace{.4pt}20064}}}


\bibitem{liu2016towards}
M.~Liu, J.~Shi, Z.~Li, C.~Li, J.~Zhu, and S.~Liu.
\newblock Towards better analysis of deep convolutional neural networks.
\newblock {\em IEEE Transactions on Visualization and Computer Graphics}, 23(1):91--100, 2017. \href{https://doi.org/10.1109/TVCG.2016.2598831}
{doi: {{%
10\hspace{.1pt}\discretionary{.}{%
}{.}\hspace{.4pt}1109\discretionary{/}{%
}{/}TVCG\hspace{.1pt}\discretionary{.}{%
}{.}\hspace{.4pt}2016\hspace{.1pt}\discretionary{.}{%
}{.}\hspace{.4pt}2598831}}}


\bibitem{liu2019interative}
S.~Liu, C.~Chen, Y.~Lu, F.~Ouyang, and B.~Wang.
\newblock An interactive method to improve crowdsourced annotations.
\newblock {\em IEEE Transactions on Visualization and Computer Graphics}, 25(1):235--245, 2019. \href{https://doi.org/10.1109/TVCG.2018.2864843}
{doi: {{%
10\hspace{.1pt}\discretionary{.}{%
}{.}\hspace{.4pt}1109\discretionary{/}{%
}{/}TVCG\hspace{.1pt}\discretionary{.}{%
}{.}\hspace{.4pt}2018\hspace{.1pt}\discretionary{.}{%
}{.}\hspace{.4pt}2864843}}}


\bibitem{Liu_Liang_Gitter_2019}
S.~Liu, Y.~Liang, and A.~Gitter.
\newblock Loss-balanced task weighting to reduce negative transfer in multi-task learning.
\newblock In {\em Proceedings of the AAAI Conference on Artificial Intelligence}, pp. 9977--9978, 2019. \href{https://doi.org/10.1609/aaai.v33i01.33019977}
{doi: {{%
10\hspace{.1pt}\discretionary{.}{%
}{.}\hspace{.4pt}1609\discretionary{/}{%
}{/}aaai\hspace{.1pt}\discretionary{.}{%
}{.}\hspace{.4pt}v33i01\hspace{.1pt}\discretionary{.}{%
}{.}\hspace{.4pt}33019977}}}


\bibitem{liu2016ssd}
W.~Liu, D.~Anguelov, D.~Erhan, C.~Szegedy, S.~Reed, C.-Y. Fu, and A.~C. Berg.
\newblock {SSD}: Single shot {MultiBox} detector.
\newblock In {\em Proceedings of the European Conference on Computer Vision}, pp. 21--37, 2016.

\bibitem{loper2002nltk}
E.~Loper and S.~Bird.
\newblock {NLTK}: The natural language toolkit.
\newblock {\em arxiv preprint arXiv:cs/0205028}, 2002.

\bibitem{lyu2024supercomputer}
F.~Lyu, C.~Chen, J.~Zhang, X.~Feng, and Z.~Tang.
\newblock Visualization for supercomputer system: A survey.
\newblock {\em Journal of Computer-Aided Design and Computer Graphics}, 36(3):321--335, 2024. \href{https://doi.org/10.3724/SP.J.1089.2024.2023-00791}
{doi: {{%
10\hspace{.1pt}\discretionary{.}{%
}{.}\hspace{.4pt}3724\discretionary{/}{%
}{/}SP\hspace{.1pt}\discretionary{.}{%
}{.}\hspace{.4pt}J\hspace{.1pt}\discretionary{.}{%
}{.}\hspace{.4pt}1089\hspace{.1pt}\discretionary{.}{%
}{.}\hspace{.4pt}2024\hspace{.1pt}\discretionary{.}{%
}{.}\hspace{.4pt}2023\discretionary{%
}{-}{-}00791}}}


\bibitem{mcinnes2018umap}
L.~McInnes, J.~Healy, and J.~Melville.
\newblock {UMap}: Uniform manifold approximation and projection for dimension reduction.
\newblock {\em arXiv preprint arXiv:1802.03426}, 2018.

\bibitem{micallef2017towards}
L.~Micallef, G.~Palmas, A.~Oulasvirta, and T.~Weinkauf.
\newblock Towards perceptual optimization of the visual design of scatterplots.
\newblock {\em IEEE Transactions on Visualization and Computer Graphics}, 23(6):1588--1599, 2017. \href{https://doi.org/10.1109/tvcg.2017.2674978}
{doi: {{%
10\hspace{.1pt}\discretionary{.}{%
}{.}\hspace{.4pt}1109\discretionary{/}{%
}{/}tvcg\hspace{.1pt}\discretionary{.}{%
}{.}\hspace{.4pt}2017\hspace{.1pt}\discretionary{.}{%
}{.}\hspace{.4pt}2674978}}}


\bibitem{openai2023gpt4}
OpenAI.
\newblock {GPT-4} technical report.
\newblock {\em arXiv, preprint arXiv: 2303.08774}, 2023.

\bibitem{parkhi2012pets}
O.~Parkhi, A.~Vedaldi, A.~Zisserman, and C.~Jawahar.
\newblock Cats and dogs.
\newblock In {\em Proceedings of the Conference on Computer Vision and Pattern Recognition}, pp. 3498--3505, 2012.

\bibitem{radford2021learning}
A.~Radford, J.~W. Kim, C.~Hallacy, A.~Ramesh, G.~Goh, S.~Agarwal, G.~Sastry, A.~Askell, P.~Mishkin, J.~Clark, G.~Krueger, and I.~Sutskever.
\newblock Learning transferable visual models from natural language supervision.
\newblock In {\em Proceedings of the International Conference on Machine Learning}, vol. 139, pp. 8748--8763, 2021.

\bibitem{reddy2018survey}
C.~K. Reddy and B.~Vinzamuri.
\newblock A survey of partitional and hierarchical clustering algorithms.
\newblock In {\em Data clustering}, pp. 87--110. Chapman and Hall/CRC, 2018. \href{https://doi.org/10.1201/9781315373515-4}
{doi: {{%
10\hspace{.1pt}\discretionary{.}{%
}{.}\hspace{.4pt}1201\discretionary{/}{%
}{/}9781315373515\discretionary{%
}{-}{-}4}}}


\bibitem{rombach2022high}
R.~Rombach, A.~Blattmann, D.~Lorenz, P.~Esser, and B.~Ommer.
\newblock High-resolution image synthesis with latent diffusion models.
\newblock In {\em Proceedings of the IEEE/CVF Conference on Computer Vision and Pattern Recognition}, pp. 10684--10695, 2022. \href{https://doi.org/10.1109/cvpr52688.2022.01042}
{doi: {{%
10\hspace{.1pt}\discretionary{.}{%
}{.}\hspace{.4pt}1109\discretionary{/}{%
}{/}cvpr52688\hspace{.1pt}\discretionary{.}{%
}{.}\hspace{.4pt}2022\hspace{.1pt}\discretionary{.}{%
}{.}\hspace{.4pt}01042}}}


\bibitem{shi2016diversifying}
L.~Shi and Y.-D. Shen.
\newblock Diversifying convex transductive experimental design for active learning.
\newblock In {\em Proceedings of the International Joint Conference on Artificial Intelligence}, pp. 1997--2003, 2016.

\bibitem{tang2006analysis}
E.~K. Tang, P.~N. Suganthan, and X.~Yao.
\newblock An analysis of diversity measures.
\newblock {\em Machine Learning}, 65(1):247–271, 2006. \href{https://doi.org/10.1007/s10994-006-9449-2}
{doi: {{%
10\hspace{.1pt}\discretionary{.}{%
}{.}\hspace{.4pt}1007\discretionary{/}{%
}{/}s10994\discretionary{%
}{-}{-}006\discretionary{%
}{-}{-}9449\discretionary{%
}{-}{-}2}}}


\bibitem{trabucco2023effective}
B.~Trabucco, K.~Doherty, M.~Gurinas, and R.~Salakhutdinov.
\newblock Effective data augmentation with diffusion models.
\newblock In {\em Proceedings of the International Conference on Learning Representations}, 2024.

\bibitem{van2008tsne}
L.~Van~der Maaten and G.~Hinton.
\newblock Visualizing data using t-sne.
\newblock {\em Journal of Machine Learning Research}, 9(11):2579--2605, 2008.

\bibitem{wang2023reprompt}
Y.~Wang, S.~Shen, and B.~Y. Lim.
\newblock {RePrompt}: Automatic prompt editing to refine ai-generative art towards precise expressions.
\newblock In {\em Proceedings of the CHI Conference on Human Factors in Computing Systems}, pp. 23--32, 2023. \href{https://doi.org/10.1145/3544548.3581402}
{doi: {{%
10\hspace{.1pt}\discretionary{.}{%
}{.}\hspace{.4pt}1145\discretionary{/}{%
}{/}3544548\hspace{.1pt}\discretionary{.}{%
}{.}\hspace{.4pt}3581402}}}


\bibitem{wang2024promptcharm}
Z.~Wang, Y.~Huang, D.~Song, L.~Ma, and T.~Zhang.
\newblock {PromptCharm}: Text-to-image generation through multi-modal prompting and refinement.
\newblock In {\em Proceedings of the CHI Conference on Human Factors in Computing Systems}, pp. 1--21, 2024.

\bibitem{wen2024convergence}
Z.~Wen.
\newblock Convergence of end-to-end training in deep unsupervised contrastive learning.
\newblock {\em arXiv preprint arXiv:2002.06979}, 2024.

\bibitem{wold1987pca}
S.~Wold, K.~Esbensen, and P.~Geladi.
\newblock Principal component analysis.
\newblock {\em Chemometrics and Intelligent Laboratory Systems}, 2(1-3):37--52, 1987. \href{https://doi.org/10.1007/springerreference_84147}
{doi: {{%
10\hspace{.1pt}\discretionary{.}{%
}{.}\hspace{.4pt}1007\discretionary{/}{%
}{/}springerreference\_84147}}}


\bibitem{xia2022cdr}
J.~Xia, L.~Huang, W.~Lin, X.~Zhao, J.~Wu, Y.~Chen, Y.~Zhao, and W.~Chen.
\newblock Interactive visual cluster analysis by contrastive dimensionality reduction.
\newblock {\em IEEE Transactions on Visualization and Computer Graphics}, 29(1):734--744, 2022. \href{https://doi.org/10.1109/tvcg.2022.3209423}
{doi: {{%
10\hspace{.1pt}\discretionary{.}{%
}{.}\hspace{.4pt}1109\discretionary{/}{%
}{/}tvcg\hspace{.1pt}\discretionary{.}{%
}{.}\hspace{.4pt}2022\hspace{.1pt}\discretionary{.}{%
}{.}\hspace{.4pt}3209423}}}


\bibitem{xiang2019interactive}
S.~Xiang, X.~Ye, J.~Xia, J.~Wu, Y.~Chen, and S.~Liu.
\newblock Interactive correction of mislabeled training data.
\newblock In {\em 2019 IEEE Conference on Visual Analytics Science and Technology}, pp. 57--68, 2019. \href{https://doi.org/10.1109/vast47406.2019.8986943}
{doi: {{%
10\hspace{.1pt}\discretionary{.}{%
}{.}\hspace{.4pt}1109\discretionary{/}{%
}{/}vast47406\hspace{.1pt}\discretionary{.}{%
}{.}\hspace{.4pt}2019\hspace{.1pt}\discretionary{.}{%
}{.}\hspace{.4pt}8986943}}}


\bibitem{xie2019semantic}
X.~Xie, X.~Cai, J.~Zhou, N.~Cao, and Y.~Wu.
\newblock A semantic-based method for visualizing large image collections.
\newblock {\em IEEE Transactions on Visualization and Computer Graphics}, 25(7):2362--2377, 2019. \href{https://doi.org/10.1109/TVCG.2018.2835485}
{doi: {{%
10\hspace{.1pt}\discretionary{.}{%
}{.}\hspace{.4pt}1109\discretionary{/}{%
}{/}TVCG\hspace{.1pt}\discretionary{.}{%
}{.}\hspace{.4pt}2018\hspace{.1pt}\discretionary{.}{%
}{.}\hspace{.4pt}2835485}}}


\bibitem{xu2021dash}
Y.~Xu, L.~Shang, J.~Ye, Q.~Qian, Y.-F. Li, B.~Sun, H.~Li, and R.~Jin.
\newblock Dash: Semi-supervised learning with dynamic thresholding.
\newblock In {\em International Conference on Machine Learning}, pp. 11525--11536, 2021.

\bibitem{yamaguchi2023on}
S.~Yamaguchi and T.~Fukuda.
\newblock On the limitation of diffusion models for synthesizing training datasets.
\newblock In {\em NeurIPS Workshop on Synthetic Data Generation with Generative AI}, 2023.

\bibitem{yang2023foundation}
W.~Yang, M.~Liu, Z.~Wang, and S.~Liu.
\newblock Foundation models meet visualizations: Challenges and opportunities.
\newblock {\em Computational Visual Media}, 2023.
\newblock to be published. \href{https://doi.org/10.1007/s41095-023-0393-x}
{doi: {{%
10\hspace{.1pt}\discretionary{.}{%
}{.}\hspace{.4pt}1007\discretionary{/}{%
}{/}s41095\discretionary{%
}{-}{-}023\discretionary{%
}{-}{-}0393\discretionary{%
}{-}{-}x}}}


\bibitem{yang2022diagnosing}
W.~Yang, X.~Ye, X.~Zhang, L.~Xiao, J.~Xia, Z.~Wang, J.~Zhu, H.~Pfister, and S.~Liu.
\newblock Diagnosing ensemble few-shot classifiers.
\newblock {\em IEEE Transactions on Visualization and Computer Graphics}, 28(9):3292--3306, 2022. \href{https://doi.org/10.1109/TVCG.2022.3182488}
{doi: {{%
10\hspace{.1pt}\discretionary{.}{%
}{.}\hspace{.4pt}1109\discretionary{/}{%
}{/}TVCG\hspace{.1pt}\discretionary{.}{%
}{.}\hspace{.4pt}2022\hspace{.1pt}\discretionary{.}{%
}{.}\hspace{.4pt}3182488}}}


\bibitem{ye2024mfm}
Y.~Ye, S.~Xiao, X.~Zeng, and W.~Zeng.
\newblock {ModalChorus}: Visual probing and alignment of multi-modal embeddings via modal fusion map.
\newblock {\em arXiv preprint arXiv:2407.12315}, 2024. \href{https://doi.org/10.1109/tvcg.2024.3456387}
{doi: {{%
10\hspace{.1pt}\discretionary{.}{%
}{.}\hspace{.4pt}1109\discretionary{/}{%
}{/}tvcg\hspace{.1pt}\discretionary{.}{%
}{.}\hspace{.4pt}2024\hspace{.1pt}\discretionary{.}{%
}{.}\hspace{.4pt}3456387}}}


\bibitem{zhang2024adapt}
Q.~Zhang, C.~Chen, Z.~Liu, and Z.~Tang.
\newblock {I-Adapt}: Using iou adapter to improve pseudo labels in cross-domain object detection.
\newblock In {\em European Conference on Artificial Intelligence}, pp. 57--64, 2024.

\bibitem{zhang2023expanding}
Y.~Zhang, D.~Zhou, B.~Hooi, K.~Wang, and J.~Feng.
\newblock Expanding small-scale datasets with guided imagination.
\newblock In {\em Proceedings of the Advances in Neural Information Processing Systems}, pp. 76558--76618, 2023.

\bibitem{zhou2024cluster}
Y.~Zhou, W.~Yang, J.~Chen, C.~Chen, Z.~Shen, X.~Luo, L.~Yu, and S.~Liu.
\newblock Cluster-aware grid layout.
\newblock {\em IEEE Transactions on Visualization and Computer Graphics}, 30(1):240--250, 2024. \href{https://doi.org/10.1109/TVCG.2023.3326934}
{doi: {{%
10\hspace{.1pt}\discretionary{.}{%
}{.}\hspace{.4pt}1109\discretionary{/}{%
}{/}TVCG\hspace{.1pt}\discretionary{.}{%
}{.}\hspace{.4pt}2023\hspace{.1pt}\discretionary{.}{%
}{.}\hspace{.4pt}3326934}}}


\bibitem{zhou2019evolutionary}
Z.-H. Zhou, Y.~Yu, and C.~Qian.
\newblock {\em Evolutionary Learning: Advances in Theories and Algorithms}.
\newblock Springer Singapore, 2019. \href{https://doi.org/10.1007/978-981-13-5956-9}
{doi: {{%
10\hspace{.1pt}\discretionary{.}{%
}{.}\hspace{.4pt}1007\discretionary{/}{%
}{/}978\discretionary{%
}{-}{-}981\discretionary{%
}{-}{-}13\discretionary{%
}{-}{-}5956\discretionary{%
}{-}{-}9}}}


\bibitem{zhu2024distribution}
H.~Zhu, L.~Yang, J.-H. Yong, W.~Zhang, and B.~Wang.
\newblock Distribution-aware data expansion with diffusion models.
\newblock {\em arXiv preprint arXiv:2403.06741}, 2024.

\end{thebibliography}

\end{document}